\documentclass[10pt,twocolumn,letterpaper]{article}

\usepackage{iccv}
\usepackage{times}
\usepackage{epsfig}
\usepackage{graphicx}
\usepackage{amsmath}
\usepackage{amssymb}
\usepackage[accsupp]{axessibility}  
\usepackage{tabularx}
\usepackage{kotex}
\def\etal{\emph{et al}\onedot}
\usepackage{xcolor}

\newcommand{\Fref}[1]{Figure \ref{#1}}
\newcommand{\Sref}[1]{Section \ref{#1}}
\newcommand{\Tref}[1]{Table \ref{#1}}

\usepackage{capt-of}
\usepackage{lipsum}

\newcommand\blfootnote[1]{%
  \begingroup
  \renewcommand\thefootnote{}\footnote{#1}%
  \addtocounter{footnote}{-1}%
  \endgroup
}

\usepackage[pagebackref=true,breaklinks=true,letterpaper=true,colorlinks,bookmarks=false]{hyperref}

\iccvfinalcopy 

\def\iccvPaperID{3644} 

\ificcvfinal\fi

\begin{document}

\title{Domain-Aware Universal Style Transfer}
\author{Kibeom Hong \textsuperscript{1}\qquad Seogkyu Jeon\textsuperscript{1}\qquad Huan Yang\textsuperscript{3}\qquad Jianlong Fu\textsuperscript{3} \qquad Hyeran Byun\textsuperscript{1,2*}\\
\textsuperscript{1}Department of Computer Science, Yonsei University\\
\textsuperscript{2}Graduate school of AI, Yonsei University\\
\textsuperscript{3}Microsoft Research\\
{\tt\small \{cha2068,jone9312,hrbyun\}@yonsei.ac.kr\qquad\{huayan,jianf\}@microsoft.com}
}
\pagenumbering{gobble}
\ificcvfinal\thispagestyle{empty}\fi

\twocolumn[{
\renewcommand\twocolumn[1][]{#1}
\maketitle
\begin{center}
    \centering
    \includegraphics[width=0.98\linewidth]{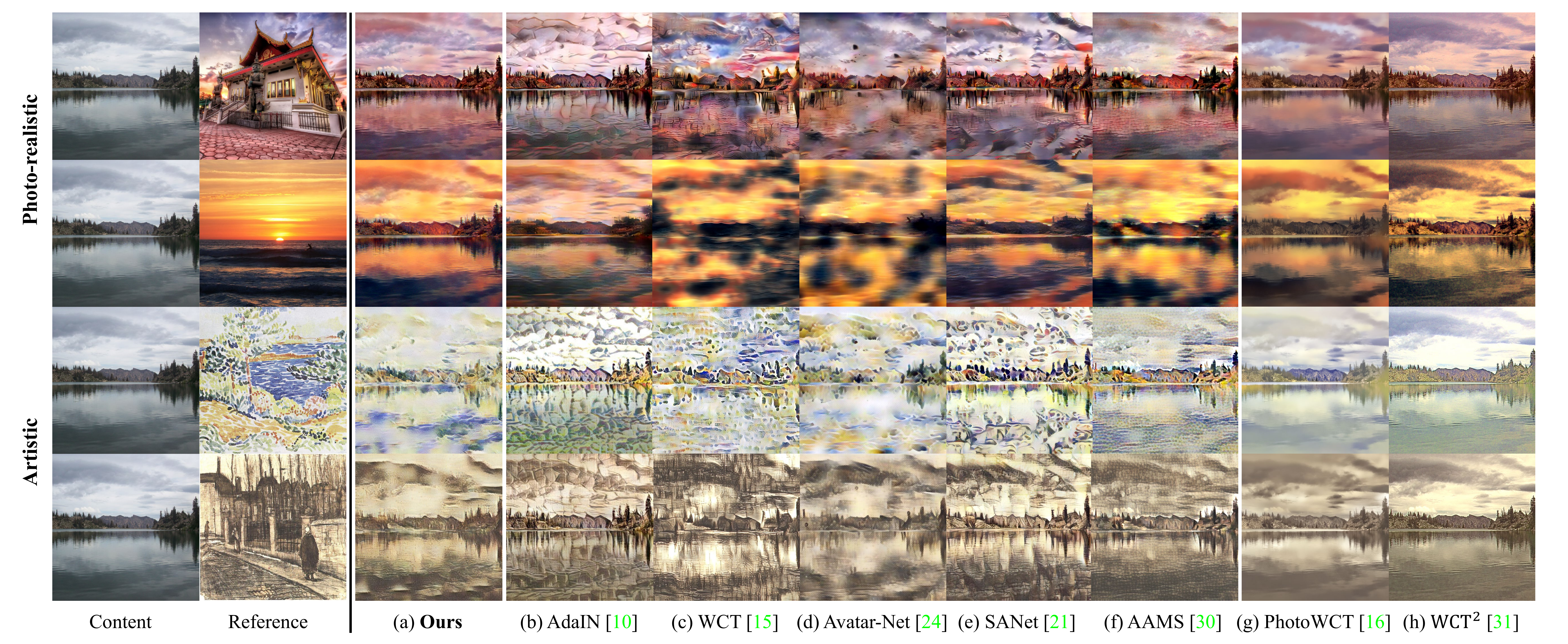}
    \captionof{figure}{Stylization results. Given input pairs~(a fixed content and four references from photo and art domains), we compare our method (a) with both artistic transfer methods~(b-f) and photo-realistic methods~(g-h). While previous works fail to generate plausible results with references coming from the opposite domains, our model successfully conducts style transfer in both domains.~(Best viewed in color.)}
    \label{fig:teaser}
\end{center}
}]
\blfootnote{* Corresponding author}

\begin{abstract}
Style transfer aims to reproduce content images with the styles from reference images. Existing universal style transfer methods successfully deliver arbitrary styles to original images either in an artistic or a photo-realistic way. However, the range of “arbitrary style” defined by existing works is bounded in the particular domain due to their structural limitation. Specifically, the degrees of content preservation and stylization are established according to a predefined target domain. As a result, both photo-realistic and artistic models have difficulty in performing the desired style transfer for the other domain. To overcome this limitation, we propose a unified architecture, \textbf{D}omain-aware \textbf{S}tyle \textbf{T}ransfer \textbf{N}etworks (DSTN) that transfer not only the style but also the property of domain (i.e., domainness) from a given reference image. To this end, we design a novel domainness indicator that captures the domainness value from the texture and structural features of reference images. Moreover, we introduce a unified framework with domain-aware skip connection to adaptively transfer the stroke and palette to the input contents guided by the domainness indicator. Our extensive experiments validate that our model produces better qualitative results and outperforms previous methods in terms of proxy metrics on both artistic and photo-realistic stylizations. All codes and pre-trained weights are available at
{\small \href{https://github.com/Kibeom-Hong/Domain-Aware-Style-Transfer}{Kibeom-Hong/Domain-Aware-Style-Transfer}}.

\end{abstract}

\begin{figure*}[t]
        \centering
        \includegraphics[width=0.975\linewidth]{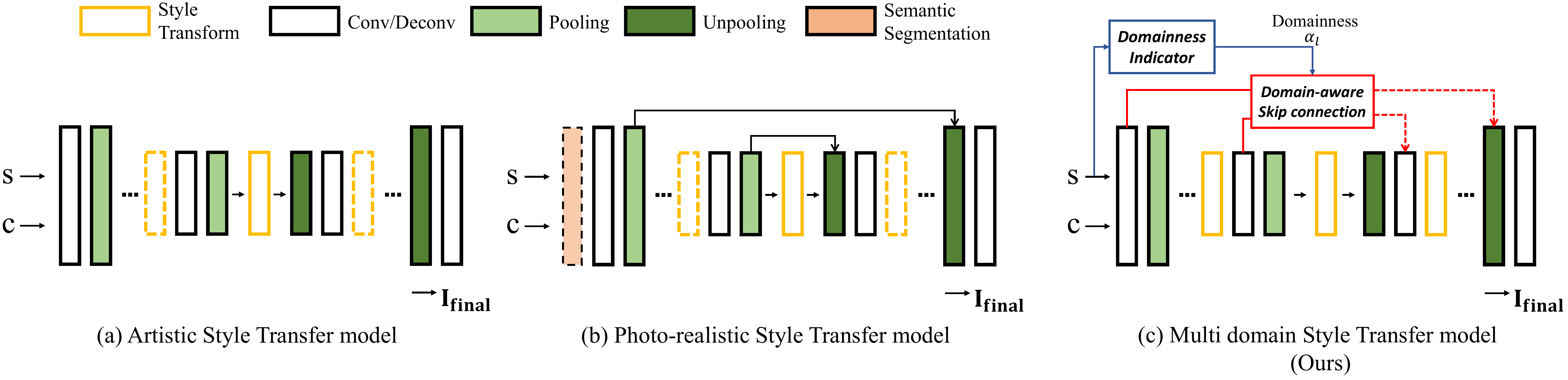}
        \captionof{figure}{Conceptual differences between previous style transfer models and our model~(DSTN).
        Artistic style transfer models~(a) adopt encoder-decoder structures to transfer the global style pattern, but struggle to preserve the original contexts of input images. With the help of the skip connection, photo-realistic style transfer models~(b) successfully preserve the structural information, but they lack the ability of transferring the delicate style. Different from the previous methods, our networks~(c) deliver the structural information of content samples adaptively based on the domainness of a given style image. The dashed line of style transformation and semantic segmentation block indicate that they can be omitted. Note that given the content~\textit{C} and style~\textit{S}, each model outputs the final image~(I$_{final}$)}
        \label{fig:abstracness}
        \vspace{-3mm}
\end{figure*}
\vspace{-6mm}
\section{Introduction}

Recreating an image with the style of another image has been a long-standing research topic. As a seminal work, Gatys~\etal~\cite{gatys2016image} propose the neural style transfer with deep features extracted from pre-trained networks,~\ie{, VGG-19 model~\cite{simonyan2014very}}. After that, thanks to a series of previous studies, we can meet several stylized pictures of various painting styles of famous painters~(\eg{, Van Gogh}).

Existing universal style transfer methods show the ability to deal with arbitrary reference images on either artistic or photo-realistic domain. On one hand, WCT~\cite{li2017universal} and AdaIN~\cite{huang2017arbitrary} transform the features of content images to match second-order statistics of reference features. Followed by these approaches, several artistic transfer studies~\cite{sheng2018avatar,Zhang2019MultimodalST,Cheng_2021_CVPR,Wang_2021_CVPR,Hu2020AestheticAwareIS} endeavor to transfer the global style pattern from reference images. Meanwhile, photo-realistic style transfer studies~\cite{luan2017deep,li2018closed,yoo2019photorealistic,An2020UltrafastPS} focus on preserving original structures while transferring target styles.

However, we observe the limitation that the meaning of “arbitrary” is restricted in a specific domain~(\ie{, either artistic or photo-realistic}), and it comes from fundamental structural modifications for predefined target domains, as shown in~\Fref{fig:abstracness}. In specific, artistic style transfer models have difficulty in maintaining clear details in the decoder because there is no clue directly coming from the content image. As a result, structural distortions of the content image occur when the artistic style transfer methods confront the photo-realistic reference images~(\Fref{fig:teaser}~(b)-(f)). On the other hand, photo-realistic style transfer models heavily constrain the transformation of input references with skip connections. Thus, they lack the ability to express delicate patterns~(\eg{, stroke pattern}) of artistic references~(\Fref{fig:teaser}~(g)-(h)). In summary, existing arbitrary style transfer models generate undesired outputs when they receive samples of the other domain as references.

To overcome this limitation, we focus on capturing the domain characteristics from a given reference image and adjust the degree of the stylization and the structural preservation adaptively. For this purpose, we propose \textit{\textbf{D}omain-aware \textbf{S}tyle \textbf{T}ransfer \textbf{N}etworks}~(DSTN) which are a unified architecture composed of an auto-encoder with domain-aware skip connections and the domainness indicator. First, we introduce the~\textit{domain-aware skip connection} to balance between content preservation and texture stylization for the domain-aware universal style transfer. Unlike the conventional skip connection which conveys intact structural details, the proposed skip connection block adjusts the transmission clarity of the high-frequency component from the stylized feature maps according to the domain properties.

To obtain the domain property~(\ie{, domainness}) from a given reference image, we design the \textit{domainness indicator}. Our novel indicator analyzes the characteristics of the domain by utilizing both the texture and structural feature maps extracted from different levels of our encoder. In order to predict a continuous domain factor with the range of $\left[ 0,1 \right]$, we propose to augment the intermediate space between art and photo with mixed samples.

With the proposed domain-aware architecture, DSTNs deliver semantic and structural information enabling artistic and photo-realistic style transfer, respectively. Consequently, DSTNs generate impeccable stylized results for arbitrary style, regardless of the target domain~(\Fref{fig:teaser}~(a)).

\begin{figure*}[t]
        \centering
        \includegraphics[width=1.0\linewidth]{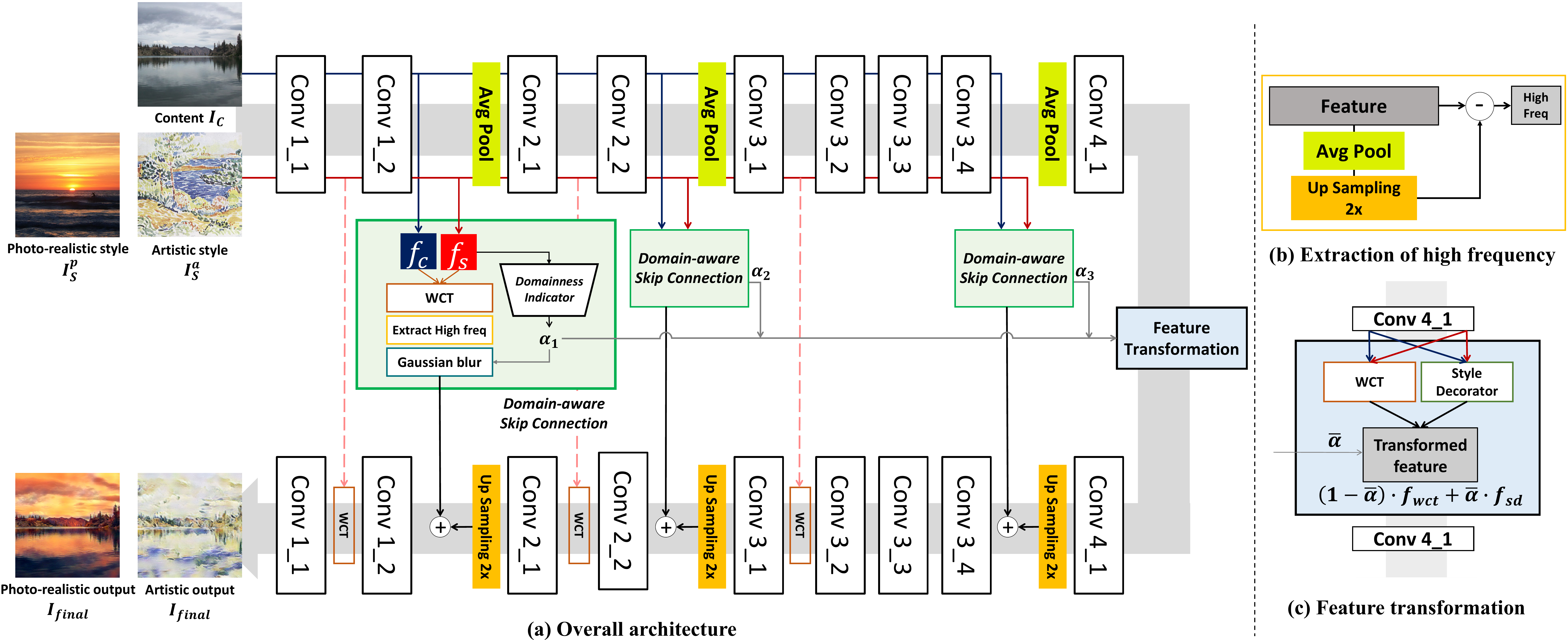}
        \captionof{figure}{Overview of the domain-aware style transfer networks~(DSTN). (b) shows the process of extracting the high-frequency components and (c) depicts the detail of feature transformation.}
        \label{fig:Arich}
\end{figure*}

In experiments, we train our decoder and the domainness indicator with Microsoft COCO~\cite{lin2014microsoft} and WikiArt~\cite{phillips2011wiki} datasets. Qualitatively, we show that DSTNs are capable of generating plausible stylized results on both domains. Besides, through quantitative proxy metrics and  user study, we demonstrate that our method outperforms previous methods on both photo-realistic and artistic style transfer.

Our contributions are three-fold: 1) We propose the novel end-to-end unified architecture, domain-aware style transfer networks~(DSTN), for multi-domain style transfer. 2) With the proposed domainness indicator and domain-aware skip connections, we capture the domain characteristics and adaptively balance between the content preservation and the texture transformation. 3) DSTNs achieve the admirable performance given references of both photo-realistic and artistic domains, and outperform previous methods in terms of preservation and stylization.

\section{Related works}
Since the pioneering work of Gatys~\etal~\cite{gatys2016image} opens up a new research area named \textit{Neural Style Transfer}, many works have explored stylization methods with the power of representation ability of neural networks.

\noindent\textbf{Artistic neural style transfer.}
Johnson~\etal~\cite{johnson2016perceptual} and Ulyanov~\etal~\cite{ulyanov2017improved} have directly trained feed-forward generative networks and achieve faster style transfer compared to image optimization methods. Still, these models need to be trained whenever they confront a new style. To alleviate this limitation, Li~\etal~\cite{li2017universal} have proposed the whitening and coloring transformation~(WCT) for arbitrary style transfer. Huang~\etal~\cite{huang2017arbitrary} introduce the adaptive instance normalization~(AdaIN) to simplify costly ZCA transformation by using the mean and standard deviation of the style features. Meanwhile, Chen~\etal~\cite{chen2016fast} and Sheng~\etal~\cite{sheng2018avatar} propose to swap the patches of content feature and normalized feature with the most correlated style features through the deconvolution methods, respectively. However, artistic transfer models tend to cause spatial distortion when generating photo-realistic styles.

\noindent\textbf{Photo-realistic neural style transfer.} 
~Since artistic neural style transfer methods have difficulty in preserving original structures of content images, many attempts for photo-realistic style transfer have been made.
Luan~\etal~\cite{luan2017deep} propose the deep photo style transfer with photorealism regularization term based on the Matting Laplacian~\cite{levin2007closed}. Li~\etal~\cite{li2018closed} replace the upsampling of the decoder with an unpooling layer and pass the index key of max pooling. In addition, they adopt additional post-processing steps,~\eg{, smoothing and filtering}. Yoo~\etal~\cite{yoo2019photorealistic} introduce the wavelet transform to preserve structural details. Yet, these models have difficulty in expressing textures of art paintings since they heavily constrain the level of stylization.

\noindent\textbf{Multi-domain neural style transfer.}
A few previous studies are capable of conducting style transfer on both artistic and photo-realistic domains.
Li~\etal~\cite{li2019learning} propose the linear style transfer~(LST) with the learnable linear transformation matrix for universal style transfer. They additionally adopt the spatial propagation network~\cite{liu2017learning} for photo-realistic style transfer. Chiu~\etal~\cite{Chiu2020Iterative} introduce an iterative transformation for stylized features with analytical gradient descent. However, these two models require additional information during inference,~\ie{, the domain of a given reference}, and they should adopt the auxiliary module or determine the number of iterative steps. In contrast, our DSTNs capture the domain-related characteristics from given reference inputs and achieve the multi-domain style transfer without any extra guidance.

\section{Method}
In this section, we describe the proposed domain-aware style transfer networks~(DSTN) in detail. DSTNs consist of an auto-encoder with \textit{domain-aware skip connections} and the \textit{domainness indicator}. Notably, our goal is transferring the style of an arbitrary reference image from artistic~($I^{a}_{S}$) or photo-realistic~($I^{p}_{S}$) domain to a content image~($I_{C}$). 

The overview of our method is illustrated in~\Fref{fig:Arich}~(a). We first describe the proposed auto-encoder with domain-aware skip connection for multi-domain style transfer. Thereafter, we detail the proposed domainness indicator which captures the property of domain from a given reference image.

\subsection{Domain-aware Skip Connection}
As discussed in earlier sections, the skip-connection is a fundamental difference that separates photo-realistic methods from artistic ones. It is advantageous in terms of structural preservation, but it constrains delicate artistic expressions. Based on this observation, we propose a domain-aware skip connection to conduct domain-aware universal style transfer on both photo-realistic and artistic domains. The domain-aware skip connection transforms the content feature $f_c$ with the given reference feature $f_s$. Thereafter, we extract the high-frequency components of a stylized feature as shown in~\Fref{fig:Arich}~(b). We exploit this high-frequency components as a key of reconstruction, since it contains the structural information of $I_c$. Consequently, the decoder reconstructs an image with structural information coming from skip connections and texture information from the feature transformation block. 

With this architecture design, we can adjust the level of structural preservation according to the domain properties of reference images. For artistic references, we deliberately blur the high-frequency components, so that our decoder reconstructs the image relying on deep texture features rather than structural details. We contort the high-frequency information with the Gaussian kernel of $\sigma = 16$ with the kernel size of $\left \lfloor \alpha_{l}\times8 \right \rfloor + 1$. The  $\alpha_{l}$ indicates the domain property obtained from the proposed domainness indicator which is detailed in the following section. As the blur increases in proportion to kernel size, a reference image with high $\alpha_{l}$ results in artistic stylization. In the opposite case of low $\alpha_{l}$, the decoder utilizes the clear high-frequency components, thus resulting in photo-realistic results. Through ablation studies in section~\ref{experiment:ablation study}, we verify the effectiveness of each component in the proposed domain-aware skip connection.

\begin{figure}[t]
        \centering
        \includegraphics[width=1.0\linewidth]{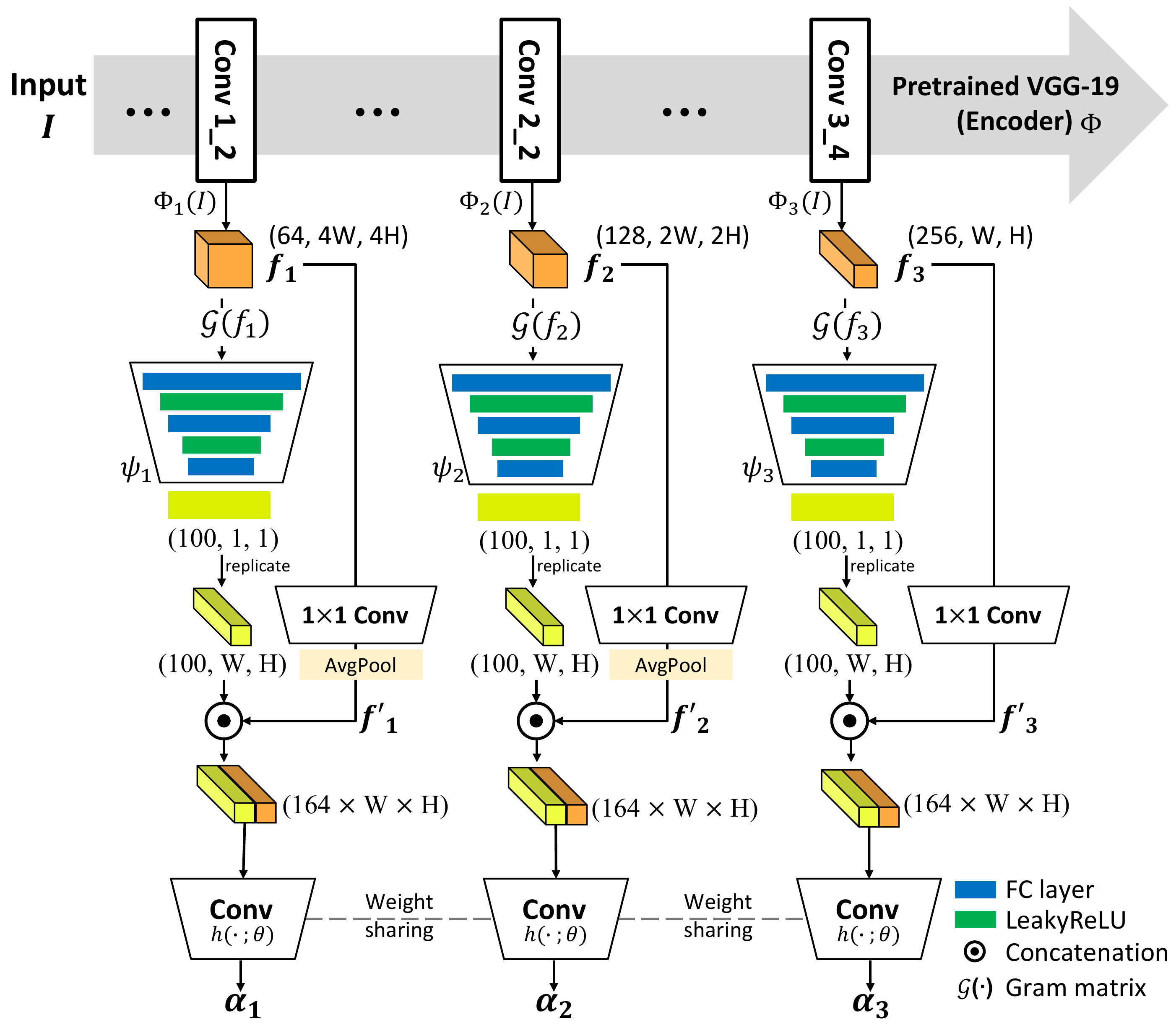}
        \captionof{figure}{Overview of the proposed domainness indicator. It is designed to analyze the domainness based on multi-scale feature representations from VGG layers. We share the weights of the last convolutional layers for multi-level features.}
        \label{fig:styleindicator}
\end{figure}

\subsection{Domainness Indicator}

To capture the domain property~(\ie{, domainness}), we introduce the domainness indicator which exploits structural features as well as textural features. As shown in~\Fref{fig:styleindicator}, our indicator takes feature maps~$f_{l}$ from three different layers~$\Phi_{l}$~(\ie{,~{\fontfamily{qcr}\selectfont Conv1\_2, Conv2\_2} and {\fontfamily{qcr}\selectfont Conv3\_4}}) of our encoder. Afterwards, we obtain the texture information by calculating a gram matrix~\cite{gatys2016image}~($\mathcal{G}(f_{l})$) of feature maps $f_{l}$. In addition, to encode structural information, we apply the channel-wise pooling on $f_{l}$ through the 1$\times$1 Conv layer. Lastly, we concatenate the texture and the structural information, which are forwarded through the weight-shared convolutional layers $h$ to obtain the domainness $\alpha_{l}$ of each level~$l$. These steps are formulated as:
\begin{equation}
\label{eq1:extract features}
\begin{split}
    f_{l} &= \Phi_{l}(I),~l\in\left\{1,2,3\right\}, \\
    F^{DI}_{l}(I) &= h~([\psi_{l}((\mathcal{G}(f_{l})))~\odot~f'_{l}]), \\
    \alpha_{l} &= \sigma(F^{DI}_{l}(I)),
\end{split}
\end{equation}
where $\odot$ denotes the channel-wise concatenation operator, $\psi$ is fully-connected layers, $f'_{l}$ is the channel-wise pooled feature maps of each level, and $\sigma(\cdot)$ is the sigmoid operation. We adopt the spatial average pooling for $f_{1}$ and ${f_{2}}$ to ensure the spatial sizes of features from different levels.

Furthermore, we adopt a domain adaptation method~\cite{gong2019dlow} to utilize the intermediate space between photo-realistic and artistic. We define the intermediate samples as follows:
\begin{equation}
    I^{mix} = Mix(I^{p}, I^{a}, \beta),
\end{equation}
where $I^{p}$, $I^{a}$ and $I^{mix}$ represent photo-realistic, artistic and mixed samples respectively. $Mix$ is the Mixup~\cite{zhang2018mixup} method and $\beta$ is the strength of interpolation from the beta distribution as in~\cite{gong2019dlow}.

To make the proposed indicator learn domainness, we adopt the binary cross entropy which is formulated as:
\begin{equation}
\begin{split}
    \mathcal{L}_{bce} = \frac{1}{|L|}\sum_{l\in L}(~\mathbb{E}_{I^{a}} \left [log(F^{DI}_{l}(I^{a}))\right ] ~~~~~~~~~~~~~~~~~~~~~~~~~~~~\\
    ~~~~~~~~~~~~~~~~~~~~+ \mathbb{E}_{I^{p}} \left [log(1 - F^{DI}_{l}(I^{p}))\right ]~),
\end{split}
\end{equation}
where $L$ depicts aforementioned three layers.

For mixed samples of the intermediate domain, we train the domainness indicator to embed their features between two domains according to $\beta$:
\begin{equation}
    \begin{split}
        \mathcal{L}_{dlow} = \frac{1}{|L|}\sum_{l\in L}(~(1-&\beta)\cdot\textit{dist}(F^{DI}_{l}(I^{p}), F^{DI}_{l}(I^{mix}))\\
        &+\beta\cdot\textit{dist}(F^{DI}_{l}(I^{a}), F^{DI}_{l}(I^{mix}))~).
    \end{split}
\end{equation}
The distance between features (\textit{dist}) is calculated with $L1$ distance, as follows: 
\begin{equation}
    dist(f_{A},~f_{B})=||f_{A}-f_{B}||_{1}
\end{equation}

Lastly, we obtain our total loss function for domainness indicator as:
\begin{equation}
    \begin{split}
        \mathcal{L}_{DI} = \lambda_{bce}\mathcal{L}_{bce} + \lambda_{dlow}\mathcal{L}_{dlow},
    \end{split}
\end{equation}
where $\lambda_{bce}$ and $\lambda_{dlow}$ are weighting factors of losses.

\begin{figure*}[t]
        \centering
        \includegraphics[width=1.00\linewidth]{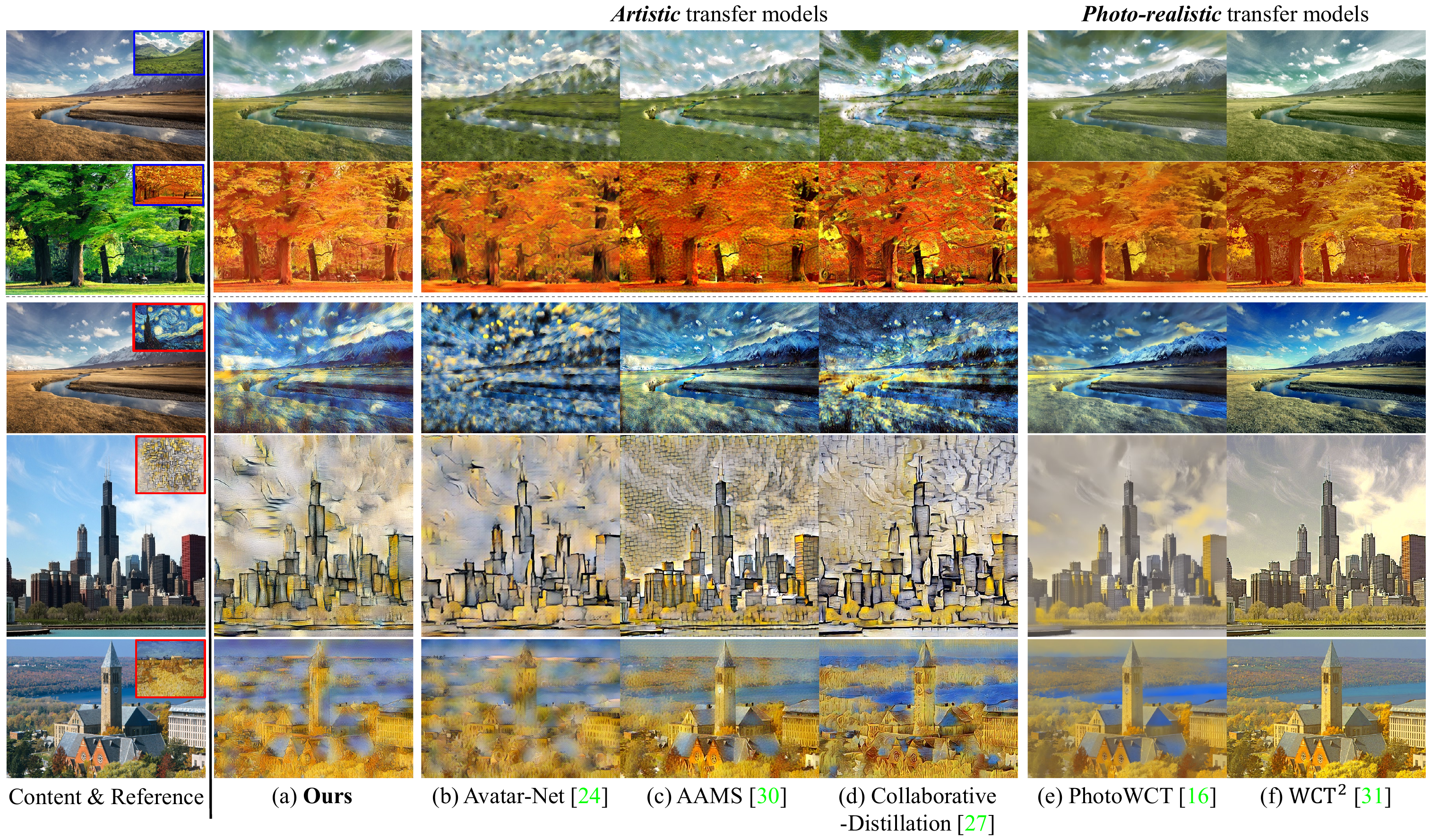}
        \captionof{figure}{Qualitative comparisons with state-of-the-art models. The \textcolor{blue}{blue box} indicates photo-realistic reference images and the \textcolor{red}{red box} indicates artistic ones. We depict the stylized results from existing artistic style transfer models~(b-d) and photo-realistic ones~(e-f). Previous models produce unsatisfactory results when they receive images from different domains, not the target of each model. The results of (a) demonstrate that our DSTNs produce both photo-realistic and artistic well regardless of the domain of given images.}
        \label{fig:qualitative results}
\end{figure*}

\begin{figure*}[t]
        \centering
        \includegraphics[width=1.0\linewidth]{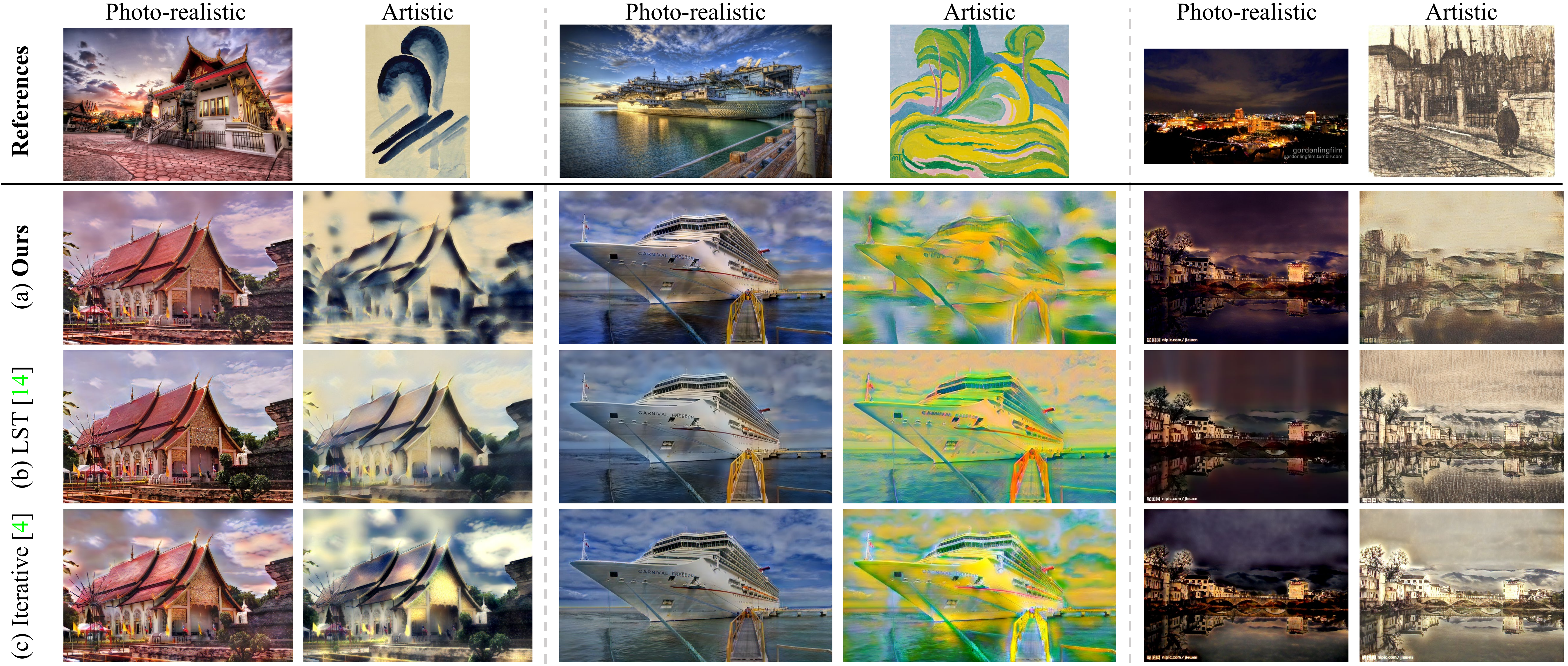}
        \captionof{figure}{Comparisons with previous state-of-the-arts for multi-domain style transfer.}
        \label{fig:qualitative hybrid results}
\end{figure*}

\subsection{Overall Architecture and Training}
We adopt a pre-trained VGG-19~(up to \texttt{conv4\_1}) as the encoder and substitute its max pooling layers with average pooling layers. The decoder mirrors the encoder and all pooling layers are replaced with up-sampling layers. Then, we set up our domain-aware skip connection between \texttt{Conv1\_2, Conv2\_2, Conv3\_4} layers of both the encoder and decoder. Inspired by Avatar-Net~\cite{sheng2018avatar}, we add short-cut connections links at \texttt{Conv1\_1, Conv2\_1, Conv3\_1} layers for better stylization.

Following previous studies~\cite{li2017universal,huang2017arbitrary,sheng2018avatar,yao2019attention,park2019arbitrary,wang2020collaborative,yang2020learning,yan2020sttn}, we apply transformation on the final feature of the encoder with the mean of $\alpha_{l}$~($\overline{\alpha}$), as shown in~\Fref{fig:Arich}~(c). In the transformation block, we control the weight of features between the universal stylized feature~\cite{li2017universal}~($f_{wct}$) which is good at preserving the global context, and the style decorated feature~\cite{sheng2018avatar}~($f_{sd}$) which represents the local style pattern well.

The decoder is trained to perform reconstruction,~\ie{, to invert the feature maps to an RGB image} with the perceptual distance~\cite{johnson2016perceptual} and the contextual similarity~\cite{mechrez2018contextual} as follows:
\begin{equation}
    \begin{split}
        \mathcal{L}_{rec} =  \lambda_{p}\sum_{k=1}^{4} \left \| \phi_{k}(I_{r}) - \phi_{k}(I_{o})  \right \|^{2}_{2} + \lambda_{cx}CX(I_{r}, I_{o}),
    \end{split}
\end{equation}
where $\phi_{k}$ denotes the activation maps after \texttt{ReLU\_$k$\_1} layers of VGG-19, $CX$ is the contextual similarity~\cite{mechrez2018contextual} between input images~($I_{o}$) and reconstructed images~($I_{r}$). $\lambda_{p}$ and $\lambda_{cx}$ indicate the hyper-parameters for weighing the perceptual distance and the contextual similarity respectively.

We replace the conventional $L2$ distance with the contextual similarity for the better quality of stylized outputs~(especially to prevent the blur issue that occurs in artistic stylization). We investigate the effect of this modification on the loss function in~\Sref{experiment:ablation study}. To further elevate the stylization performance, we exploit the multi-scale discriminator with the adversarial loss~($\mathcal{L}_{adv}$)~\cite{Goodfellow2014GenerativeAN}. The details of the multi-scale discriminator and the adversarial loss are described in the supplementary material.

Finally, we train our networks with an end-to-end manner, and the overall loss function is calculated as follows:
\begin{equation}
    \begin{split}
        \mathcal{L}_{total} = \mathcal{L}_{DI} + \mathcal{L}_{rec} + \lambda_{tv}\mathcal{L}_{tv} + \lambda_{adv}\mathcal{L}_{adv},
    \end{split}
\end{equation}
where $\mathcal{L}_{tv}$ represents the total variation loss for enforcing pixel-wise smoothness.

\section{Experiments}
\paragraph{Implementation details.}
We train DSTNs on MS-COCO~\cite{lin2014microsoft} and WikiART~\cite{phillips2011wiki} datasets, each containing roughly 80,000 images of real photos and artistic images respectively. We use the Adam~\cite{Kingma2015AdamAM} optimizer to train the domainness indicator as well as the decoder with a batch size of 6 and the learning rate is initially set to $1e$-4. Throughout the experiments, we set 256$\times$256 as the default image resolution. Weighting factors of loss functions are set as: $\lambda_{p}=0.1, \lambda_{cx}=1, \lambda_{tv}=1, \lambda_{dlow}=1$ , $\lambda_{adv}=0.1$ and $\lambda_{bce}=5$. Hyperparameters of $\mathcal{L}_{dlow}$ are set to the same as~\cite{gong2019dlow}. Our code is implemented with PyTorch~\cite{paszke2017automatic} and our model is trained on a single GTX 2080Ti. The code and the trained model will be publicly available online.

\subsection{Qualitative results}
We compare our networks with several state-of-the-art models qualitatively in terms of artistic and photo-realistic style transfer.
\Fref{fig:qualitative results} shows the stylized results of photo-realistic~(row 1,2) and artistic~(row 3,4,5) references respectively. AvatarNet~\cite{sheng2018avatar}, AAMS~\cite{yao2019attention}, and Collaborative Distillation~\cite{wang2020collaborative} show good artistic stylized outputs, but exhibit some distortions on photographic references~(\eg{, the detail of leaves or skyline of mountains}). On the other hand, PhotoWCT~\cite{li2018closed} and WCT$^{2}$~\cite{yoo2019photorealistic} successfully generate photo-realistic outputs, but they fail to express specific painting styles~(\eg{, brush pattern or detailed texture}) of artistic references.
In other words, previous methods stylize content images according to the predefined target domain without considering the domain characteristics of the input references.

On the other hand, our DSTNs show plausible results on both domains~(\Fref{fig:qualitative results}~(a)). These samples demonstrate that the domainness indicator extracts the domain property well from given reference samples, and the decoder inverts features to images adaptively according to the given domainness. Concretely, DSTNs conduct the artistic style transfer while preserving semantic information via the domain-aware skip connection rather than costly attention modules~\cite{yao2019attention, park2019arbitrary}.
Moreover, structural details delivered from the proposed skip connection enables both intact content representation and stylization in photo-realistic style transfer. We note that our networks do not require any pre-processing or post-processing step.

Furthermore, we compare our method with previous multi-domain neural style transfer models~\cite{li2019learning,Chiu2020Iterative} as shown in \Fref{fig:qualitative hybrid results}. We observe that DSTNs express not only the texture of the reference image but also the domain characteristics well.~(More  qualitative comparisons are included in our supplementary materials.)

\subsection{Quantitative results}
\begin{figure}[t]
        \centering
        \includegraphics[width=1.0\linewidth]{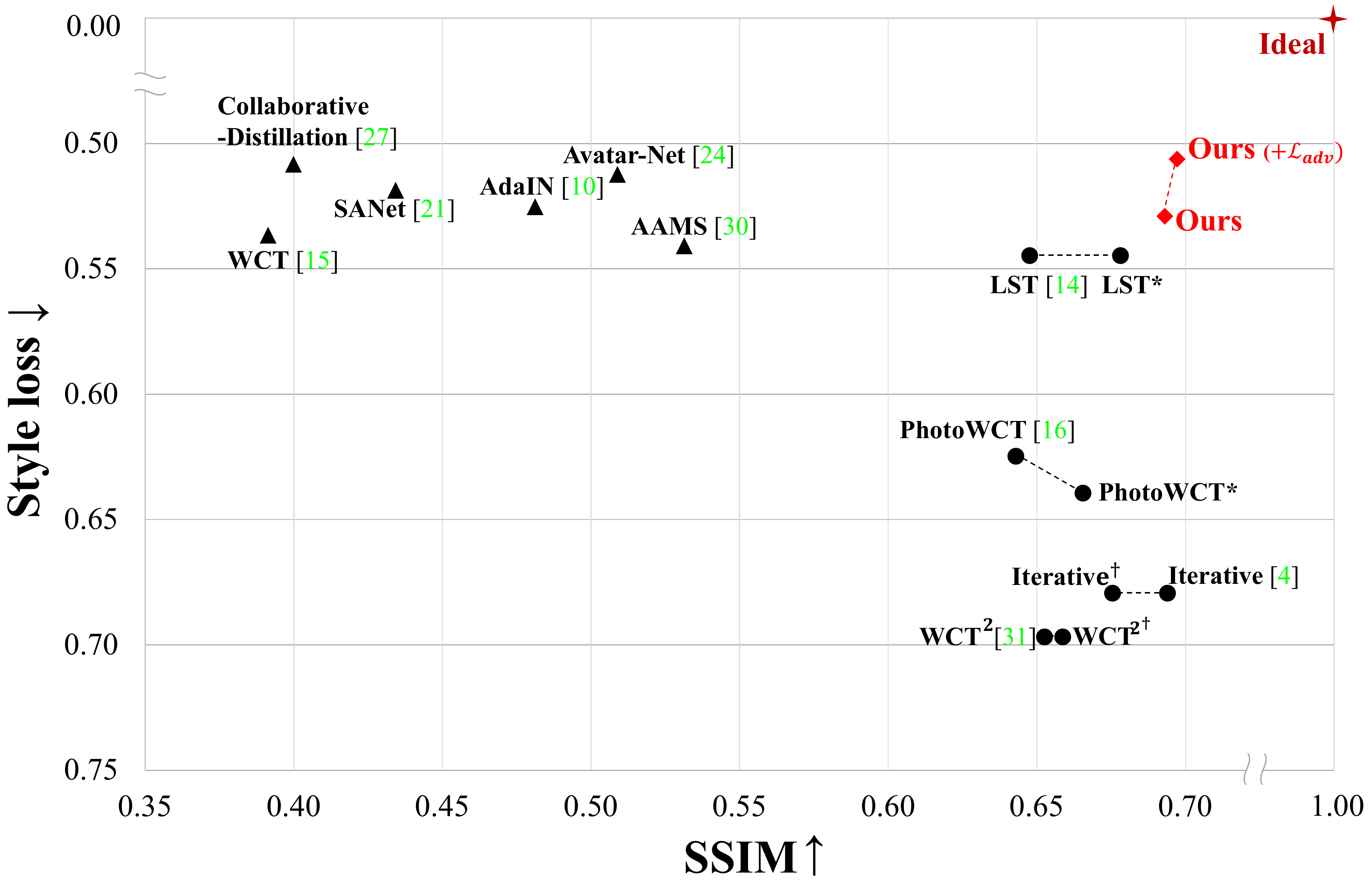}
        \captionof{figure}{Quantitative results. SSIM index~(higher is better) versus Style loss~(lower is better). Ideal case for multi-domain style transfer is the top-right corner~(red star). Dashed lines depict the gap before and after the post-processing steps or additional information,~\eg{, smoothing step~(*) or segmentation maps~($\dagger$)}.}
        \label{fig:quantitative results}
\end{figure}

\noindent\textbf{Statistics.}
To measure both photorealism and the degree of style representation, we employ two proxy metrics for structural preservation and stylization.~\Fref{fig:quantitative results} shows the plot of SSIM~(X-axis) versus style loss~(Y-axis) as quantitative results. We calculate the structural similarity~(SSIM) index between edge responses of content images and \textbf{photo-realistic} stylized outputs following WCT$^{2}$~\cite{yoo2019photorealistic}. In addition, we compute the style loss~\cite{gatys2016image} between \textbf{artistic} stylized outputs and reference samples over five crop patches with 128$\times$128 resolution. We use the 60 pairs of content and photo-realistic references provided by DPST~\cite{luan2017deep} for quantitative evaluation. Also, we sample 60 artistic references randomly from the test split of WikiART dataset.

As shown in \Fref{fig:quantitative results}, artistic style transfer models~(left top) achieve decent style loss scores but they fail to preserve original structures with los SSIM values. On the other hand, photo-realistic transfer models~(bottom right) conserve the original structure information well, but have trouble in expressing the style of a reference image. Though LST~\cite{li2019learning} achieves the good performance on both sides, we note that they require an additional spatial propagation network~(SPN)~\cite{liu2017learning} as well as domain information of references for transfer. In contrast, our DSTNs conduct the style transfer without any pre-processing~(\eg{, determining domains of given reference images or semantic segmentation}) or post-processing~(\eg{, smoothing and filtering}), and outperform other methods including the multi-domain models~\cite{li2019learning, Chiu2020Iterative} through a single training. Remarkably, our DSTNs accomplish the state-of-the-art performance on both domains with or even without the adversarial loss~($\mathcal{L}_{adv}$).

\begin{table}[t]
\begin{center}
\resizebox{1.0\linewidth}{!}
{
\begin{tabular}{lccc}
\\
\cline{1-4}\cline{1-4}
Methods & Preservation$^{p}$ & Stylization$^{a}$ & Domainness$^{o}$ \\
\hline\hline
Avatar-Net~\cite{sheng2018avatar} & 0.76\%           & 19.66\%                     &16.54\%\\
AAMS~\cite{yao2019attention} & 5.30\%           & 18.59\%                     & 20.51\%\\
Colla-Distil~\cite{wang2020collaborative} & 1.52\%           & 26.07\%                     & 19.62\%\\
DSTN~(\textit{ours}) & \textbf{92.42}\%           &\textbf{35.68}\%                     & \textbf{43.33}\%\\
\cline{1-4}

Photo-WCT~\cite{li2018closed} & 13.26\%           & 13.37\%                     & 14.87\%\\
WCT$^{2}$~\cite{yoo2019photorealistic} & 39.01\%           & 6.01\%                     & 20.64\%\\
DSTN~(\textit{ours}) & \textbf{47.73}\%           & \textbf{80.62\%}                     & \textbf{64.49}\%\\
\cline{1-4}

LST~\cite{li2019learning} & \textbf{43.94}\%           & 27.71\%                     & 29.74\%\\
Iterative~\cite{Chiu2020Iterative} & 26.89\%           & 9.88\%                     & 19.49\%\\
DSTN~(\textit{ours}) & 29.17\%           & \textbf{62.41}\%                     & \textbf{50.77}\%\\
\cline{1-4}\cline{1-4}
\end{tabular}}
\end{center}
\caption{The results of user studies. We ask the participants three questions of preference: 1) preservation of content with photo-realistic references (Preservation$^{p}$), 2) expression of style pattern with artistic references (Stylization$^{a}$) and 3) looks most similar to the domain of both photo-realistic and artistic references~(Domainness$^{o}$).}
\label{tab:user study table}
\end{table}

\begin{table}[t]
\vspace{-2mm}
\begin{center}
\resizebox{1.0\linewidth}{!}
{
\begin{tabular}{lccc}
\\
\cline{1-4}\cline{1-4}
Methods & 256$\times$256 & 512$\times$512 & 1024$\times$1024 \\
\hline\hline
AdaIN~\cite{huang2017arbitrary} &0.227&0.224&0.220\\
WCT~\cite{li2017universal} &0.640 &1.415&4.111\\
Avatar-Net~\cite{sheng2018avatar} & 0.678& 0.853&1.758\\
\cline{1-4}

Photo-WCT~\cite{li2018closed} & 2.579  & 10.870 & OOM\\
WCT$^{2}$~\cite{yoo2019photorealistic} & 0.414& 0.466&0.773\\
\cline{1-4}

LST~\cite{li2019learning} &0.151&0.213&0.540\\
Iterative~\cite{Chiu2020Iterative} &0.104&0.274&1.049\\
\cline{1-4}
DSTN~(\textit{ours}) & 0.195&0.396&1.441\\
\cline{1-4}\cline{1-4}
\end{tabular}
}
\end{center}
\caption{Execution time analysis in seconds. We compare under different resolutions. OOM denotes the out-of-memory error.}
\label{tab:calculation time}
\end{table}

\noindent\textbf{User study.}
As stylization is quite subjective, we conduct user study to better evaluate our method in terms of both photo-realistic and artistic criteria. We use 24 pairs of content and reference images, and compare our DSTNs to representative models of each domain,~\ie{, artistic~\cite{sheng2018avatar,yao2019attention,wang2020collaborative} and photo-realistic~\cite{li2018closed,yoo2019photorealistic}}. Besides, as competitors, we include two previous models~\cite{li2019learning,Chiu2020Iterative} that enable multi-domain style transfer. The stylized results are provided to the users in random order with content and style images. In user study, we ask the participants questions about the preservation of content, the expression of texture and domainness. Consequently, we collect 2340 responses from 65 subjects and the results are shown in~\Tref{tab:user study table}. Overall, it can be noticed that most of the participants favor our DSTNs over all evaluated methods.

\noindent\textbf{Execution time analysis.}
\Tref{tab:calculation time} shows the execution time comparison with other methods under different resolutions. Results are estimated with a single NVIDIA RTX 2080Ti 11GB. Since DSTNs can transfer styles of both photo and artistic domains, we conduct 60 transfers for each domain and report the average time of total 120 transfers. Regardless of the resolutions, our method achieves the comparable execution time.

\subsection{Ablation studies}
\label{experiment:ablation study}
In this section, we conduct several ablation studies on the domainness, domain-aware skip connections and reconstruction losses.

\begin{figure}[t]
        \centering
        \includegraphics[width=1.0\linewidth]{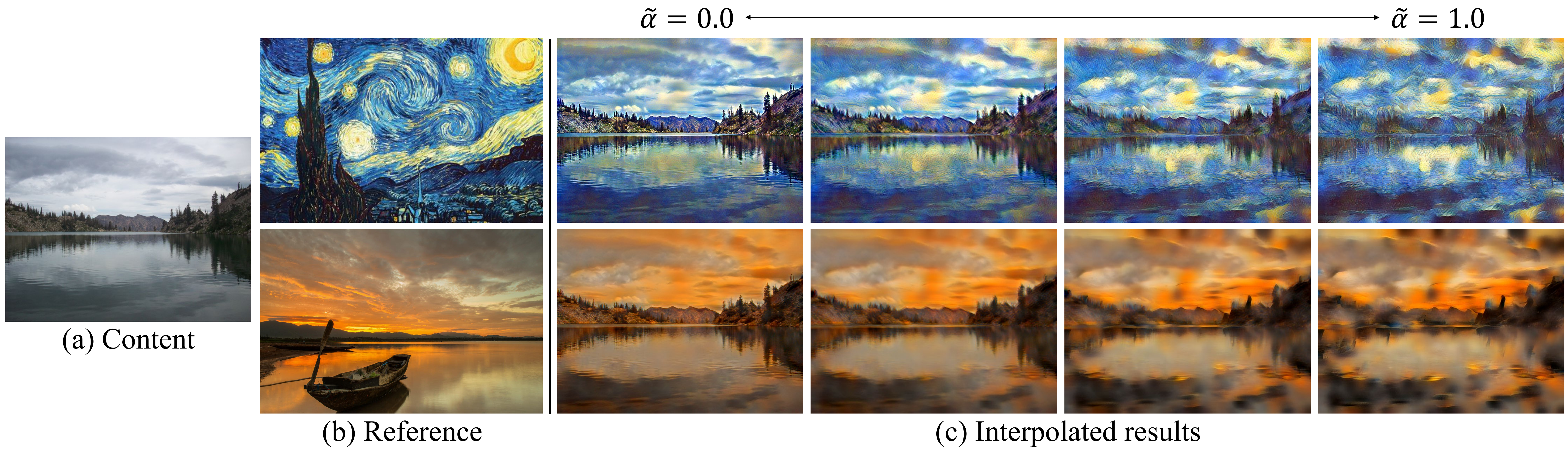}
        \captionof{figure}{Ablation study on the effect of domainness $\alpha$. More results of ablation studies are included in our supplementary materials.}
        \label{fig:ablation interpolation}
\end{figure}

\begin{figure}[t]
        \centering
        \includegraphics[width=1.0\linewidth]{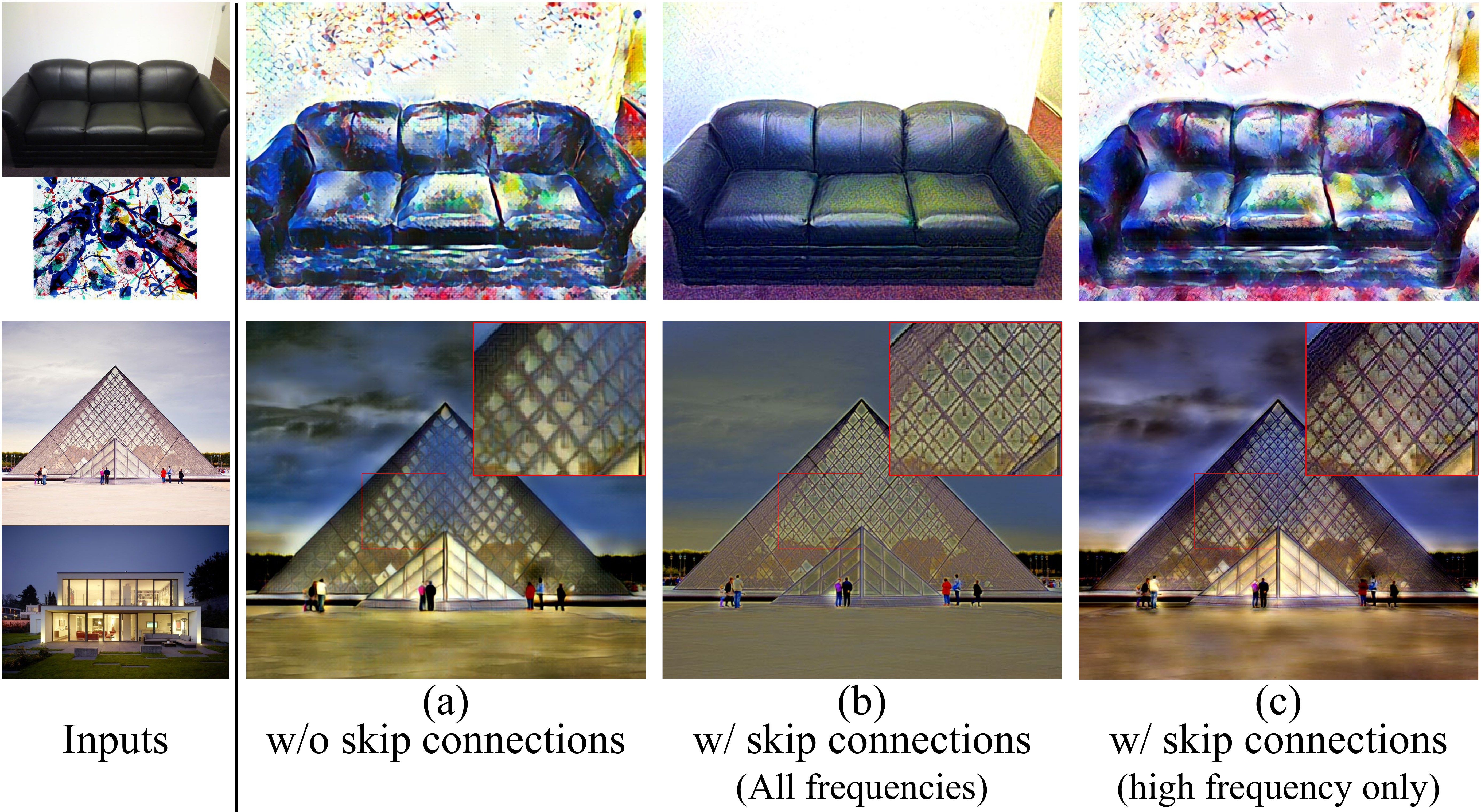}
        \captionof{figure}{Ablation study of domain-aware skip connections. We compare each case with artistic and photo-realistic references. We also display the zoomed patch in the red box for better evaluation on photo-realism.}
        \label{fig:ablation study skip}
\end{figure}

\noindent\textbf{Effect of changing the domainness $\alpha$.}
To analyze the relation between the domainness $\alpha$ and the stylized output, we intentionally set the $\alpha_{l}$ of every layer with the same value $\tilde{\alpha}$ between the range of $\left[ 0,1 \right ]$. As shown in~\Fref{fig:ablation interpolation}, outputs change smoothly between artistic and photo-realistic stylizations. Noticeably, DSTNs control the level of stylization and handle intermediate domains with a continuous domainness value.

\noindent\textbf{Analysis on domain-aware skip connections.}
Our skip connections pass the high-frequency component of content images from the encoder to the decoder with the adaptive domainness value of reference images. In~\Fref{fig:ablation study skip}, we qualitatively conduct ablation studies on domain-aware skip connections and high-frequency components. Without domain-aware skip connections, DSTNs fail to preserve structural details of the original image on photo-realistic style transfer, e.g., window frames in~\Fref{fig:ablation study skip}~(a). Though establishing domain-aware skip connections on entire encoded features enables preserving the structure of original features, degrees of transformation on artistic references are insignificant since encoded features contain too abundant features for reconstruction as in~\Fref{fig:ablation study skip}~(b). By passing and transforming high-frequency components only, we found a sweet-spot between the content preservation and the stylization as in~\Fref{fig:ablation study skip}~(c).

\begin{figure}[t]
        \centering
        \includegraphics[width=1.0\linewidth]{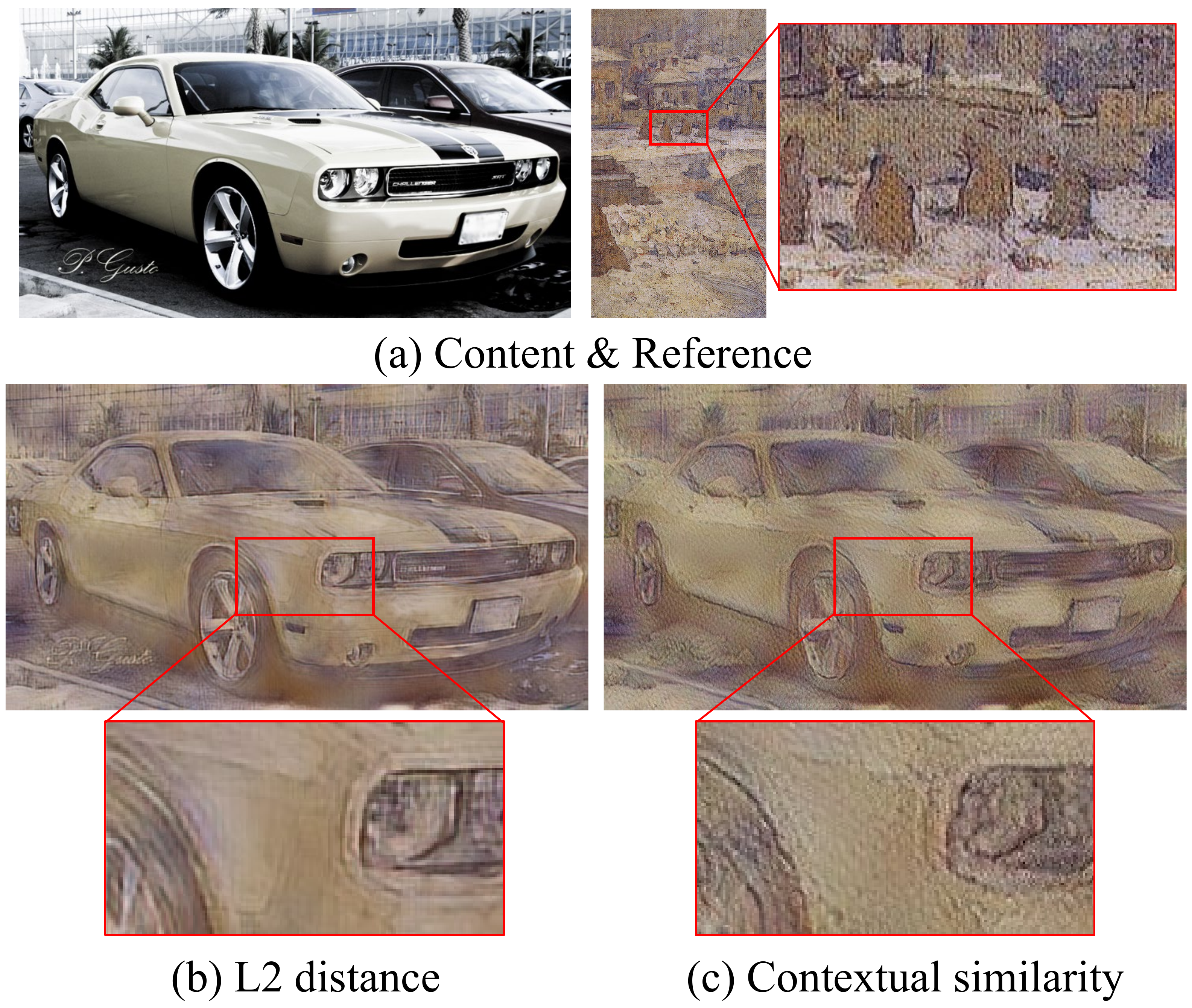}
        \captionof{figure}{Ablation study of the reconstruction loss, $\mathcal{L}_{rec}$.}
        \label{fig:ablation study loss}
\end{figure}

\noindent\textbf{Analysis of reconstruction loss.}
In DSTNs, the amount of structural clues coming from high-frequency components is less for artistic images, thus their reconstruction results get blurry. Therefore, we employ the contextual similarity~\cite{mechrez2018contextual} instead of pixel-wise $L2$ distance for reconstruction. As shown in~\Fref{fig:ablation study loss}~(b), the stylized outputs from networks trained with $L2$ distance are blurry and fail to express the detailed texture of the reference images. Contrarily, the proposed decoder trained with contextual similarity~($CX$) transfers the texture of oil painting on canvas effectively~(\Fref{fig:ablation study loss}~(c)).

\section{Conclusion}
In this work, we proposed domain-aware style transfer networks~(DSTN) for the domain-aware universal style transfer. We introduced the novel domainness indicator which captures the domain characteristics~(\ie{, domainness}) from the arbitrary references. Moreover, our domain-aware skip connection adjusts the clarity of transferred information by Gaussian blur in accordance with the domainness. As a result, DSTN generted admirable stylized outputs on both domains without any additional user-guidance. Qualitative and quantitative experiments verified that DSTNs achieve the state-of-the-art stylization performance in both photo-realistic and artistic domains without any pre-processing or post-processing.

\section{Acknowledgements}
{\small This project was partly supported by the National Research Foundation of Korea grant funded by the Korea government (MSIT) (No. 2019R1A2C2003760) and the Institute for Information \& Communications Technology Planning \& Evaluation (IITP) grant funded by the Korea government (No. 2020-0-01361: Artificial Intelligence Graduate School Program (YONSEI UNIVERSITY)). We sincerely appreciate all participants for the user study.}

{\small
\bibliographystyle{ieee_fullname}
\bibliography{final}

\begin{thebibliography}{10}\itemsep=-1pt

\bibitem{An2020UltrafastPS}
Jie An, Haoyi Xiong, Jun Huan, and Jiebo Luo.
\newblock Ultrafast photorealistic style transfer via neural architecture
  search.
\newblock {\em AAAI}, abs/1912.02398, 2020.

\bibitem{chen2016fast}
Tian~Qi Chen and Mark Schmidt.
\newblock Fast patch-based style transfer of arbitrary style.
\newblock {\em arXiv preprint arXiv:1612.04337}, 2016.

\bibitem{Cheng_2021_CVPR}
Jiaxin Cheng, Ayush Jaiswal, Yue Wu, Pradeep Natarajan, and Prem Natarajan.
\newblock Style-aware normalized loss for improving arbitrary style transfer.
\newblock In {\em CVPR}, pages 134--143, June 2021.

\bibitem{Chiu2020Iterative}
Tai-Yin Chiu and Danna Gurari.
\newblock Iterative feature transformation for fast and versatile universal
  style transfer.
\newblock In {\em ECCV}, pages 169--184, 2020.

\bibitem{gatys2016image}
Leon~A Gatys, Alexander~S Ecker, and Matthias Bethge.
\newblock Image style transfer using convolutional neural networks.
\newblock In {\em CVPR}, pages 2414--2423, 2016.

\bibitem{gong2019dlow}
Rui Gong, Wen Li, Yuhua Chen, and Luc~Van Gool.
\newblock Dlow: Domain flow for adaptation and generalization.
\newblock In {\em CVPR}, pages 2477--2486, 2019.

\bibitem{Goodfellow2014GenerativeAN}
Ian~J. Goodfellow, Jean Pouget-Abadie, M. Mirza, Bing Xu, David Warde-Farley,
  Sherjil Ozair, Aaron~C. Courville, and Yoshua Bengio.
\newblock Generative adversarial nets.
\newblock In {\em NeurIPS}, 2014.

\bibitem{zhang2018mixup}
Yann N.~Dauphin Hongyi~Zhang, Moustapha~Cisse and David Lopez-Paz.
\newblock mixup: Beyond empirical risk minimization.
\newblock {\em ICLR}, 2018.

\bibitem{Hu2020AestheticAwareIS}
Zhiyuan Hu, Jia Jia, Bei Liu, Yaohua Bu, and Jianlong Fu.
\newblock Aesthetic-aware image style transfer.
\newblock {\em ACM MM}, 2020.

\bibitem{huang2017arbitrary}
Xun Huang and Serge Belongie.
\newblock Arbitrary style transfer in real-time with adaptive instance
  normalization.
\newblock In {\em ICCV}, pages 1501--1510, 2017.

\bibitem{johnson2016perceptual}
Justin Johnson, Alexandre Alahi, and Li Fei-Fei.
\newblock Perceptual losses for real-time style transfer and super-resolution.
\newblock In {\em ECCV}, pages 694--711. Springer, 2016.

\bibitem{Kingma2015AdamAM}
Diederik~P. Kingma and Jimmy Ba.
\newblock Adam: A method for stochastic optimization.
\newblock {\em ICLR}, 2015.

\bibitem{levin2007closed}
Anat Levin, Dani Lischinski, and Yair Weiss.
\newblock A closed-form solution to natural image matting.
\newblock {\em IEEE transactions on pattern analysis and machine intelligence
  (PAMI)}, 30(2):228--242, 2007.

\bibitem{li2019learning}
Xueting Li, Sifei Liu, Jan Kautz, and Ming-Hsuan Yang.
\newblock Learning linear transformations for fast image and video style
  transfer.
\newblock In {\em CVPR}, pages 3809--3817, 2019.

\bibitem{li2017universal}
Yijun Li, Chen Fang, Jimei Yang, Zhaowen Wang, Xin Lu, and Ming-Hsuan Yang.
\newblock Universal style transfer via feature transforms.
\newblock In {\em NeurIPS}, pages 386--396, 2017.

\bibitem{li2018closed}
Yijun Li, Ming-Yu Liu, Xueting Li, Ming-Hsuan Yang, and Jan Kautz.
\newblock A closed-form solution to photorealistic image stylization.
\newblock In {\em ECCV}, pages 453--468, 2018.

\bibitem{lin2014microsoft}
Tsung-Yi Lin, Michael Maire, Serge Belongie, James Hays, Pietro Perona, Deva
  Ramanan, Piotr Doll{\'a}r, and C~Lawrence Zitnick.
\newblock Microsoft coco: Common objects in context.
\newblock In {\em ECCV}, pages 740--755. Springer, 2014.

\bibitem{liu2017learning}
Sifei Liu, Shalini De~Mello, Jinwei Gu, Guangyu Zhong, Ming-Hsuan Yang, and Jan
  Kautz.
\newblock Learning affinity via spatial propagation networks.
\newblock In {\em NeurIPS}, pages 1520--1530, 2017.

\bibitem{luan2017deep}
Fujun Luan, Sylvain Paris, Eli Shechtman, and Kavita Bala.
\newblock Deep photo style transfer.
\newblock In {\em CVPR}, pages 4990--4998, 2017.

\bibitem{mechrez2018contextual}
Roey Mechrez, Itamar Talmi, and Lihi Zelnik-Manor.
\newblock The contextual loss for image transformation with non-aligned data.
\newblock In {\em ECCV}, pages 768--783, 2018.

\bibitem{park2019arbitrary}
Dae~Young Park and Kwang~Hee Lee.
\newblock Arbitrary style transfer with style-attentional networks.
\newblock In {\em CVPR}, pages 5880--5888, 2019.

\bibitem{paszke2017automatic}
Adam Paszke, Sam Gross, Soumith Chintala, Gregory Chanan, Edward Yang, Zachary
  DeVito, Zeming Lin, Alban Desmaison, Luca Antiga, and Adam Lerer.
\newblock Automatic differentiation in {PyTorch}.
\newblock In {\em NeurIPS Autodiff Workshop}, 2017.

\bibitem{phillips2011wiki}
Fred Phillips and Brandy Mackintosh.
\newblock Wiki art gallery, inc.: A case for critical thinking.
\newblock {\em Issues in Accounting Education}, 26(3):593--608, 2011.

\bibitem{sheng2018avatar}
Lu Sheng, Ziyi Lin, Jing Shao, and Xiaogang Wang.
\newblock Avatar-net: Multi-scale zero-shot style transfer by feature
  decoration.
\newblock In {\em CVPR}, pages 8242--8250, 2018.

\bibitem{simonyan2014very}
Karen Simonyan and Andrew Zisserman.
\newblock Very deep convolutional networks for large-scale image recognition.
\newblock {\em arXiv preprint arXiv:1409.1556}, 2014.

\bibitem{ulyanov2017improved}
Dmitry Ulyanov, Andrea Vedaldi, and Victor Lempitsky.
\newblock Improved texture networks: Maximizing quality and diversity in
  feed-forward stylization and texture synthesis.
\newblock In {\em CVPR}, pages 6924--6932, 2017.

\bibitem{wang2020collaborative}
Huan Wang, Yijun Li, Yuehai Wang, Haoji Hu, and Ming-Hsuan Yang.
\newblock Collaborative distillation for ultra-resolution universal style
  transfer.
\newblock In {\em CVPR}, pages 1860--1869, 2020.

\bibitem{Wang_2021_CVPR}
Pei Wang, Yijun Li, and Nuno Vasconcelos.
\newblock Rethinking and improving the robustness of image style transfer.
\newblock In {\em CVPR}, pages 124--133, June 2021.

\bibitem{yang2020learning}
Fuzhi Yang, Huan Yang, Jianlong Fu, Hongtao Lu, and Baining Guo.
\newblock Learning texture transformer network for image super-resolution.
\newblock In {\em CVPR}, pages 5790--5799, June 2020.

\bibitem{yao2019attention}
Yuan Yao, Jianqiang Ren, Xuansong Xie, Weidong Liu, Yong-Jin Liu, and Jun Wang.
\newblock Attention-aware multi-stroke style transfer.
\newblock In {\em CVPR}, pages 1467--1475, 2019.

\bibitem{yoo2019photorealistic}
Jaejun Yoo, Youngjung Uh, Sanghyuk Chun, Byeongkyu Kang, and Jung-Woo Ha.
\newblock Photorealistic style transfer via wavelet transforms.
\newblock In {\em ICCV}, pages 9036--9045, 2019.

\bibitem{yan2020sttn}
Yanhong Zeng, Jianlong Fu, and Hongyang Chao.
\newblock Learning joint spatial-temporal transformations for video inpainting.
\newblock In {\em ECCV}, 2020.

\bibitem{Zhang2019MultimodalST}
Yulun Zhang, Chen Fang, Y. Wang, Zhaowen Wang, Zhe Lin, Yun Fu, and Jimei Yang.
\newblock Multimodal style transfer via graph cuts.
\newblock {\em ICCV}, pages 5942--5950, 2019.

\end{thebibliography}
}

\end{document}


\title{Domain-Aware Universal Style Transfer}
\author{Kibeom Hong \textsuperscript{1}\qquad Seogkyu Jeon\textsuperscript{1}\qquad Huan Yang\textsuperscript{3}\qquad Jianlong Fu\textsuperscript{3} \qquad Hyeran Byun\textsuperscript{1,2*}\\
\textsuperscript{1}Department of Computer Science, Yonsei University\\
\textsuperscript{2}Graduate school of AI, Yonsei University\\
\textsuperscript{3}Microsoft Research\\
{\tt\small \{cha2068,jone9312,hrbyun\}@yonsei.ac.kr\qquad\{huayan,jianf\}@microsoft.com}
}
\pagenumbering{gobble}


\maketitle

\section{Additional qualitative results}
\begin{figure*}[t]
        \centering
        \includegraphics[width=1.0\linewidth]{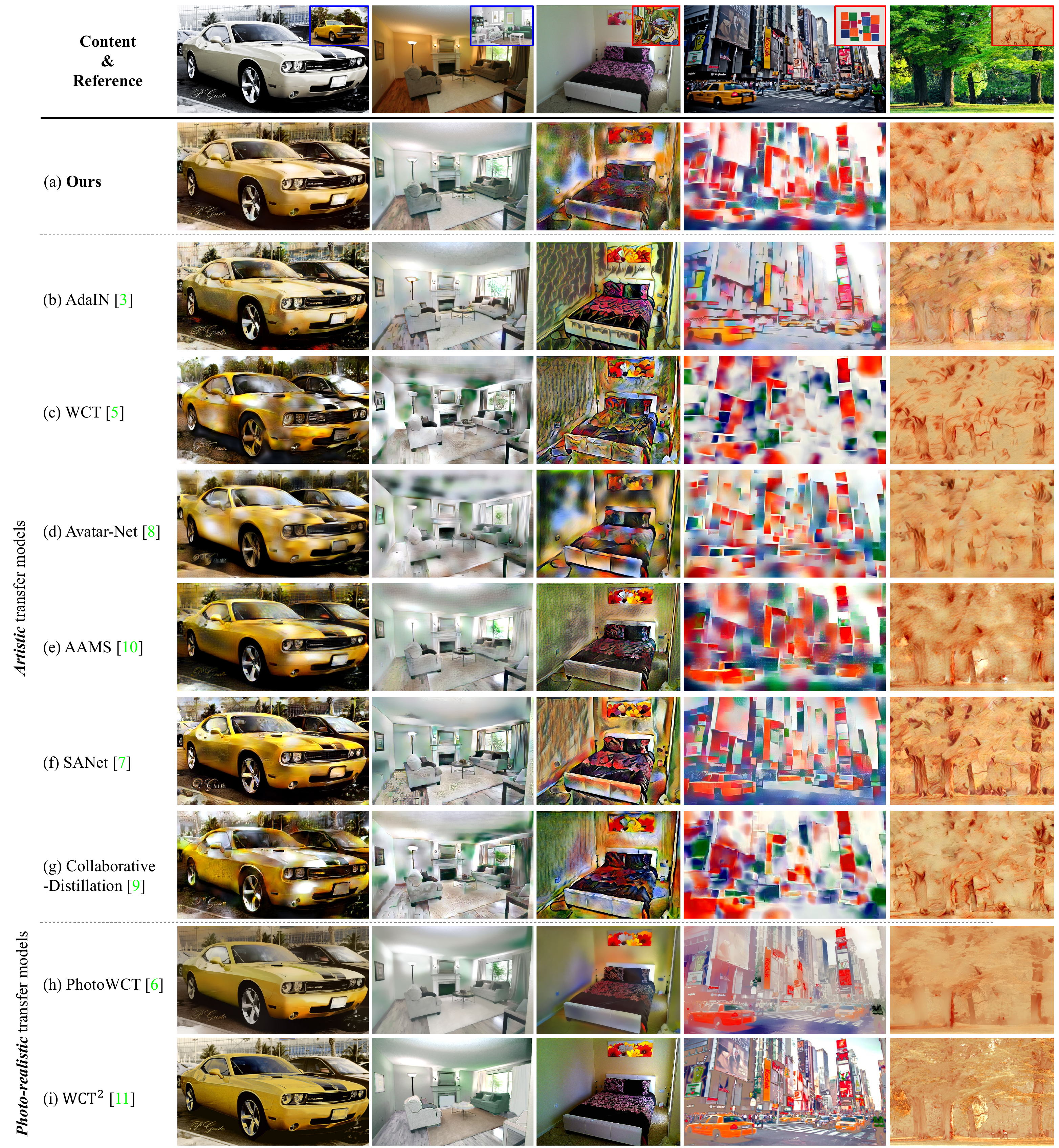}
        \captionof{figure}{Qualitative comparisons with state-of-the-art models. The \textcolor{blue}{blue box} indicates photo-realistic reference images and the \textcolor{red}{red box} indicates artistic ones. We depict the stylized results from existing artistic style transfer models~(b-g) and photo-realistic ones~(h-i). Previous methods produce unsatisfactory results when they receive images from the opposite domain. The results of (a) demonstrate that DSTNs produce both photo-realistic and artistic well regardless of the domain of reference images.}
        \label{fig:qualitative results}
\end{figure*}

\begin{figure*}[t]
        \centering
        \includegraphics[width=1.0\linewidth]{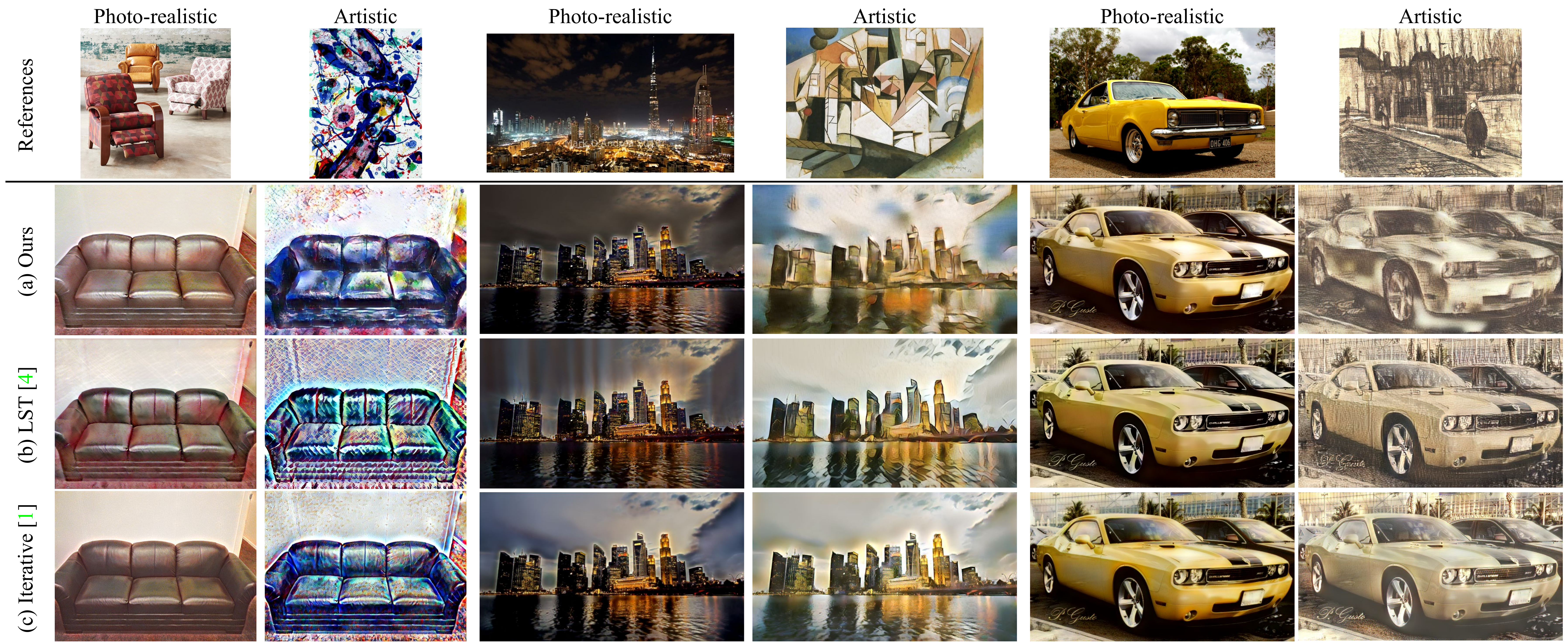}
        \captionof{figure}{Comparison with previous state-of-the-arts for multi-domain style transfer.}
        \label{fig:qualitative hybrid results}
\end{figure*}

In this section, we provide additional qualitative results for better comparison with previous state-of-the-arts. \Fref{fig:qualitative results} shows the results of single-domain methods for both artistic~\cite{huang2017arbitrary,li2017universal,park2019arbitrary,sheng2018avatar,wang2020collaborative,yao2019attention} and photo-realistic~\cite{li2018closed,yoo2019photorealistic} style transfer. In addition, \Fref{fig:qualitative hybrid results} shows the comparison with previous multi-domain style transfer methods~\cite{Chiu2020Iterative,li2019learning}. Our DSTNs outperform previous works on both domains.

\begin{figure*}[t]
        \centering
        \includegraphics[width=0.94\linewidth]{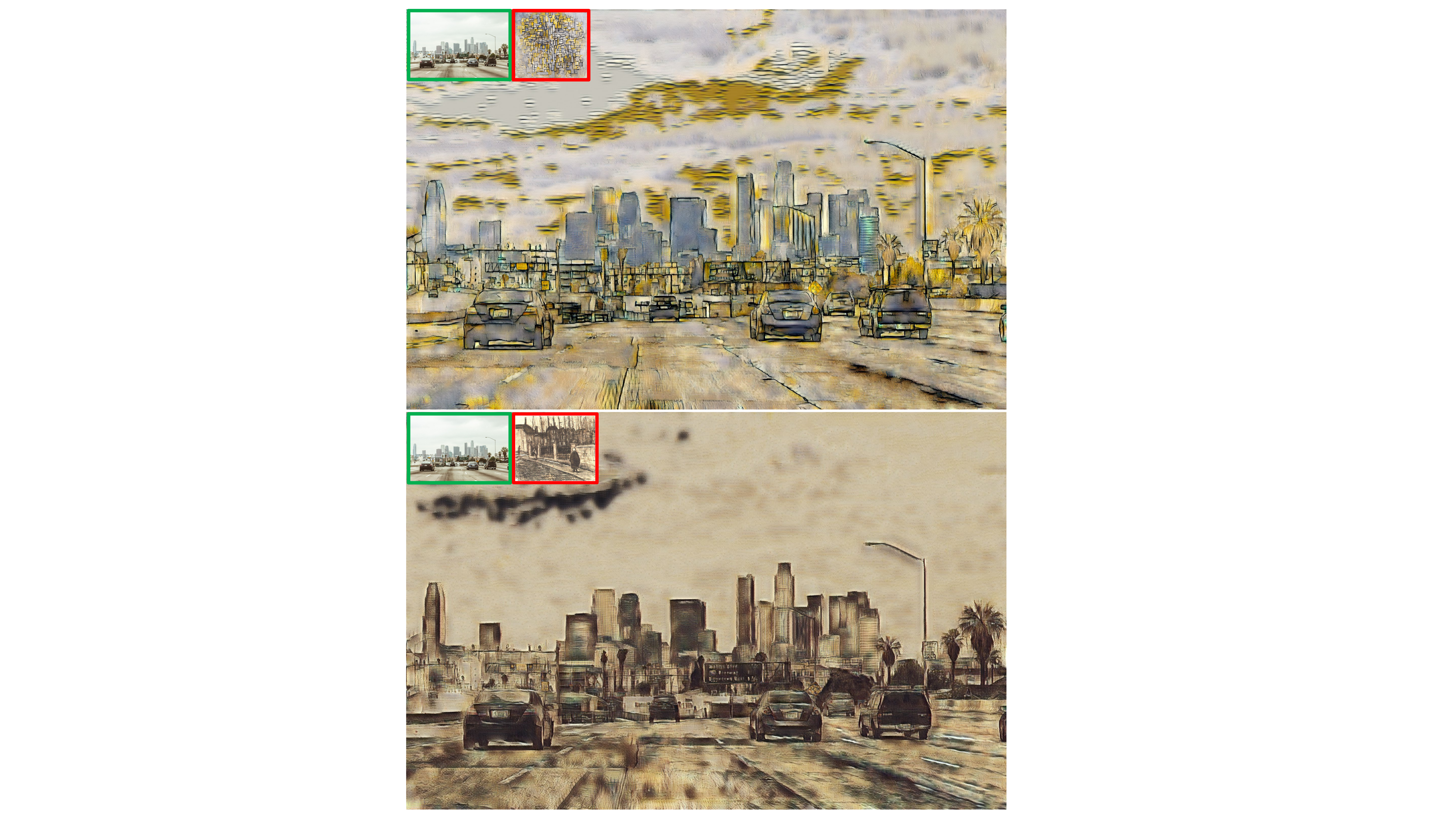}
        \captionof{figure}{Qualitative result of artistic style transfer with high resolution (4752$\times$3168). The \textcolor{red}{red box} indicates artistic reference images, and the \textcolor{darkpastelgreen}{green box} indicates the content image.}
        \label{fig:high resolution qualitative results_art}
\end{figure*}

\begin{figure*}[t]
        \centering
        \includegraphics[width=0.94\linewidth]{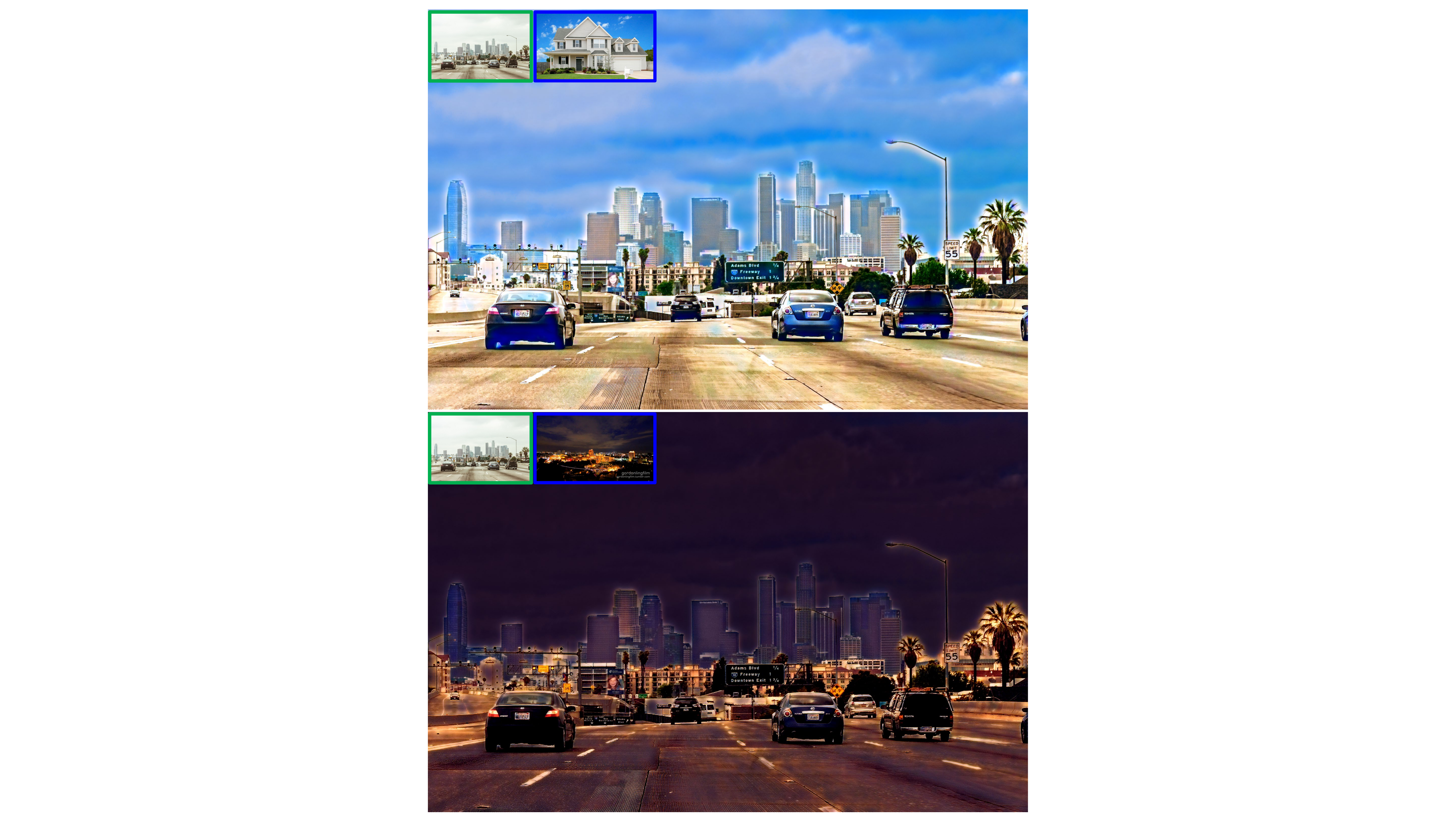}
        \captionof{figure}{Qualitative result of photo-realistic style transfer with high resolution (4752$\times$3168). The \textcolor{blue}{blue box} indicates photo-realistic reference images, and the \textcolor{darkpastelgreen}{green box} indicates the content image.}
        \label{fig:high resolution qualitative results_photo}
\end{figure*}

\begin{figure*}[t]
        \centering
        \includegraphics[width=1.0\linewidth]{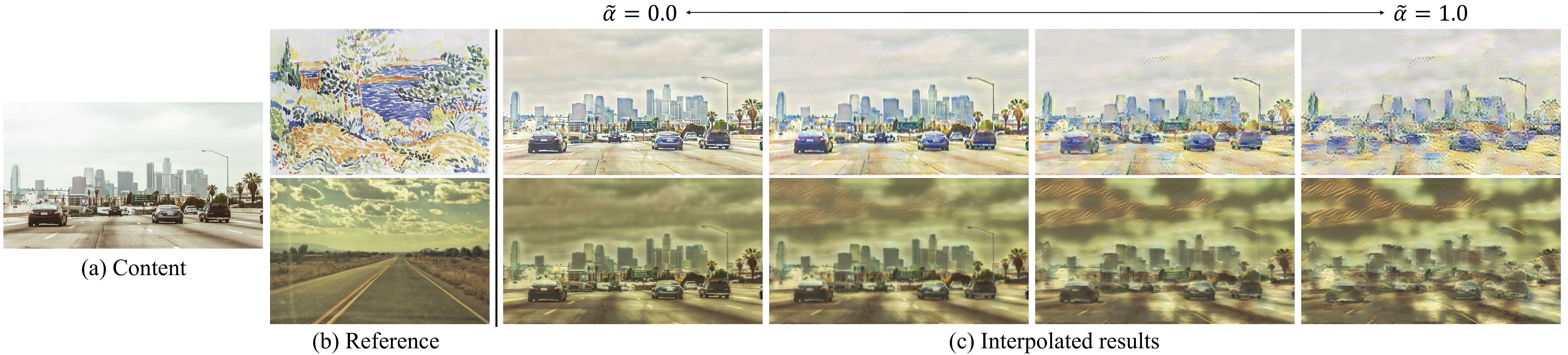}
        \captionof{figure}{Ablation study on the effect of domainness $\alpha$.}
        \label{fig:Ablation study for alpha.}
\end{figure*}

Moreover, we conduct style transfer on high-resolution images (4751 $\times$ 3168) which are shown in~\Fref{fig:high resolution qualitative results_art}, and~\Fref{fig:high resolution qualitative results_photo}. It is easy to recognize that DSTNs effectively stylize global and local patches while preserving the original semantic information. In addition, we provide additional qualitative results of ablation study on the effectiveness of domainness $\alpha$ in~\Fref{fig:Ablation study for alpha.}.


\section{Ablation study of the number of domain-aware skip connections}
\begin{figure*}[t]
        \centering
        \includegraphics[width=1.0\linewidth]{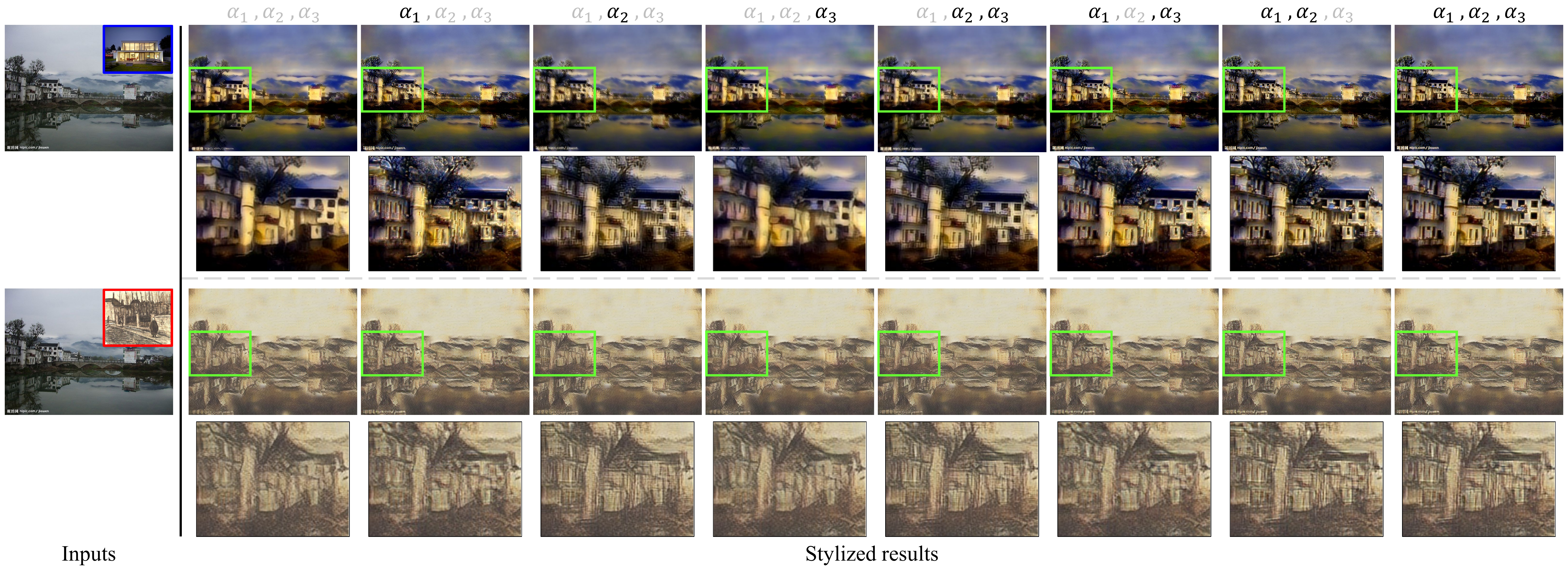}
        \captionof{figure}{Ablation study on the domain-aware skip connections. $\alpha_{l}$ denotes the domainness value from each domain-aware skip connection on the level of $l$. The gray text~(\textcolor{gray}{$\alpha_{l}$}) indicates the removal of skip connection of corresponding level. We also display the zoomed patch from the green box.}
        \label{fig:Ablation study for the number of domain-aware skip connections.}
\end{figure*}

In~\Fref{fig:Ablation study for the number of domain-aware skip connections.}, we conduct the ablation study to analyze the effect of each skip connection. As discussed in~\cite{gatys2016image}, the structural information is lost in deeper layers of the network. By fully exploiting three skip-connections, DSTNs produce the satisfactory photo-realistic stylized results. With artistic references, our domain indicator produces the higher domainness value thus the results are almost consistent.

\begin{figure*}[t]
        \centering
        \includegraphics[width=1.0\linewidth]{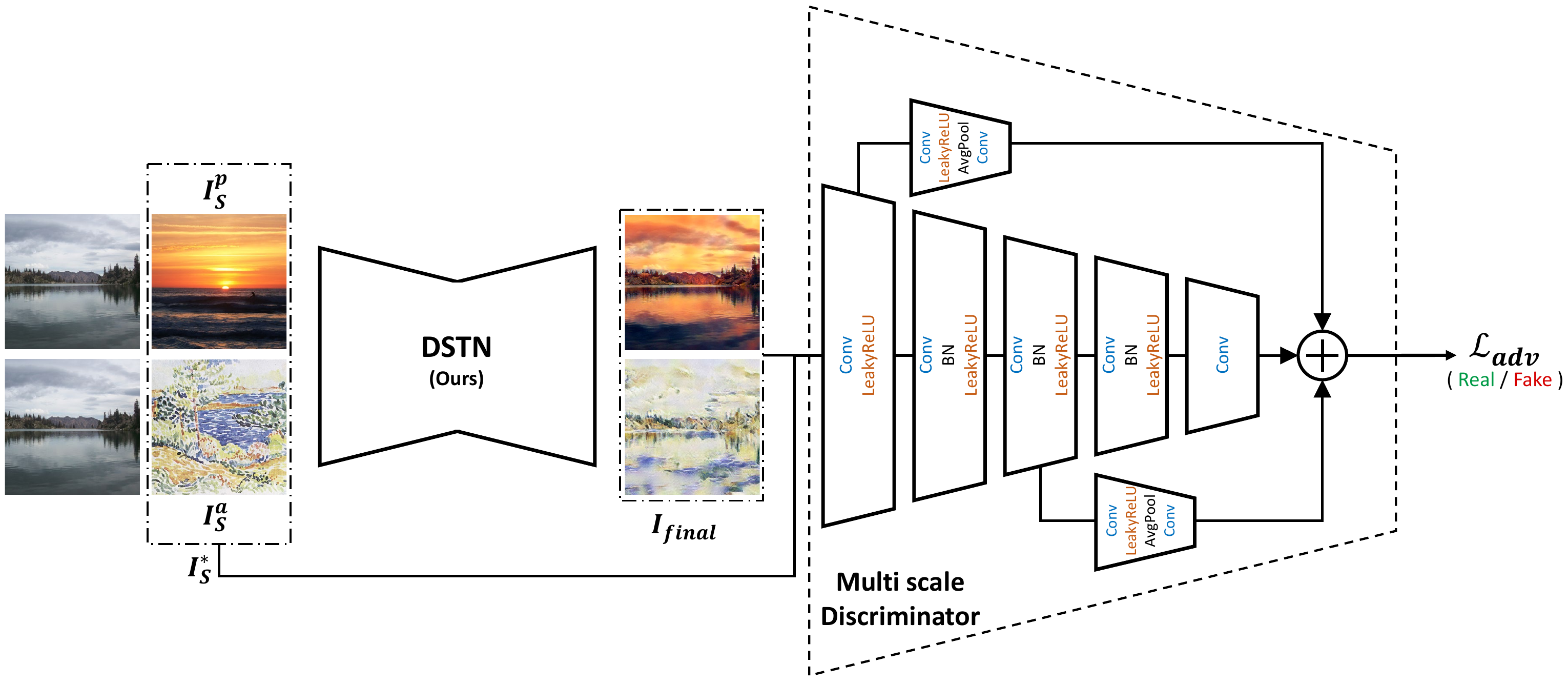}
        \captionof{figure}{Overview of the multi scale discriminator.}
        \label{fig:multi-scale discriminator}
\end{figure*}

\section{Adversarial Training via Multi-scale Discriminator}
In this section, we describe the details of the multi-scale discriminator and the adversarial training. For adversarial training, we label the original images from both datasets as \textit{real}, while stylized images are marked as \textit{fake}. We adversarially train our decoder and the discriminator so that DSTNs are capable of conducting the stylization in more realistic way.

To further enhance the performance, we adopt the multi-scale discriminator which exploits not only the global texture but also local patch-wise contexts. The multi-scale discriminator consists of four Conv blocks and one convolutional layer as illustrated in ~\Fref{fig:multi-scale discriminator}. We establish two skip-connections on the first and third Conv blocks in order to collect patch-wise features. Finally, the decoder of DSTN and the discriminator are trained with the adversarial loss as follows:
\begin{equation}
\label{eq:adverarial loss}
\begin{split}
    \mathcal{L}^{Dis}_{adv} &= \mathbb{E}_{I_{s} \sim  I^{*}_{s}}  [ \log D(I_{s}) ]  + \mathbb{E}_{\tilde{I} \sim  I_{final}}  [ \log (1- D(\tilde{I})  ]\\
    \mathcal{L}^{Dec}_{adv} &= \mathbb{E}_{\tilde{I} \sim  I_{final}}   [ \log D(\tilde{I}) ],
\end{split}
\end{equation}
where $I_{s}$ and $\tilde{I}$ denote the real images and stylized results, respectively.

\section{Photo-realistic style transfer with segmentation maps}
\begin{figure*}[t]
        \centering
        \includegraphics[width=0.8\linewidth]{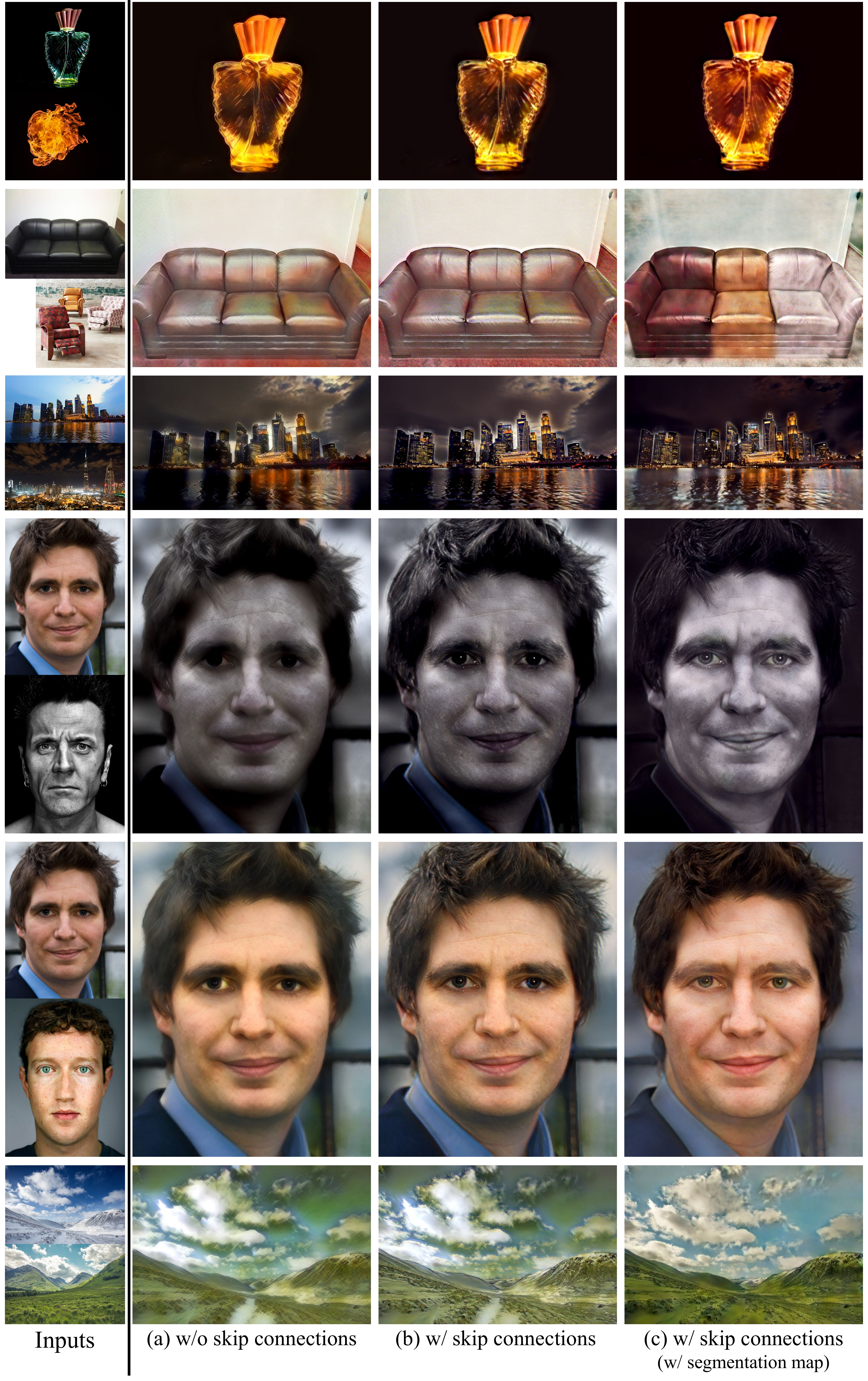}
        \captionof{figure}{Photo-realistic stylization results with segmentation maps.}
        \label{fig:qualitative segmentation results}
\end{figure*}

Following DPST~\cite{li2018closed} and WCT$^2$~\cite{yoo2019photorealistic}, DSTNs are also capable of utilizing segmentation maps to maintain semantic correspondence between content and reference images. As shown in~\Fref{fig:qualitative segmentation results}, DSTN successfully produce photo-realistic results with the help of segmentation maps. 



\clearpage

{\small
\bibliographystyle{ieee_fullname}
\bibliography{supplement}
}